\documentclass[10pt,twocolumn,letterpaper,pagenumbers]{article}

\usepackage[accsupp]{axessibility}

\usepackage[pagenumbers]{iccv} % To force page numbers, e.g. for an arXiv version

\usepackage{nopageno}
\usepackage{bbding}
\usepackage{times}
\usepackage{epsfig}
\usepackage{graphicx}
\usepackage{amsmath}
\usepackage{amssymb}
\usepackage{multirow}
\usepackage{colortbl}
\usepackage{wrapfig}
\usepackage{multido}

\definecolor{iccvblue}{rgb}{0.21,0.49,0.74}
\usepackage[pagebackref,breaklinks,colorlinks,allcolors=iccvblue]{hyperref}

\def\paperID{10093} % *** Enter the Paper ID here
\def\confName{ICCV}
\def\confYear{2025}

\title{GaussianOcc: Fully Self-supervised and Efficient 3D Occupancy Estimation with Gaussian Splatting}

\author{\textbf{$\rm Wanshui \; Gan$} $^{1, 2, *}$ \\
\and
\textbf{$\rm Fang \; Liu$} $^{1, *}$ \\
\and
\textbf{$\rm Hongbin \; Xu$} $^{3} $ \\
\and
\textbf{$\rm Ningkai \; Mo$} $^{4}$ \\
\and
\textbf{$\rm Naoto \; Yokoya$} $^{1, 2, \textsuperscript{\dag}}$ \\
\and
$\rm ^1 The \; University \; of \; Tokyo, ^2 RIKEN, \; ^3 South \; China \; University \; of \; Technology $\\
$\rm ^4 Shenzhen \; Institute \; of \; Advanced \; Technology,\ Chinese \; Academy \; of \; Sciences$ \\
$\rm ^*  \; Equal \;  contribution, \; \textsuperscript{\dag} \; Corresponding \; author $\\
{\tt\small \{wanshuigan, fangliu2896, hongbinxu1013, nk.mo19941001\}@gmail.com}\\
{\tt\small yokoya@k.u-tokyo.ac.jp}}

\begin{document}
\maketitle

\begin{abstract}

We introduce GaussianOcc, a systematic method that investigates Gaussian splatting for fully self-supervised and efficient 3D occupancy estimation in surround views. First, traditional methods for self-supervised 3D occupancy estimation still require ground truth 6D ego pose from sensors during training. To address this limitation, we propose Gaussian Splatting for Projection (GSP) module to provide accurate scale information for fully self-supervised training from adjacent view projection. Additionally, existing methods rely on volume rendering for final 3D voxel representation learning using 2D signals (depth maps and semantic maps), which is time-consuming and less effective. We propose Gaussian Splatting from Voxel space (GSV) to leverage the fast rendering properties of Gaussian splatting. As a result, the proposed GaussianOcc method enables fully self-supervised (no ground truth ego pose) 3D occupancy estimation in competitive performance with low computational cost (2.7 times faster in training and 5 times faster in rendering). The relevant code is available in \href{https://github.com/GANWANSHUI/GaussianOcc.git}{https://github.com/GANWANSHUI/GaussianOcc.git}.
 
\end{abstract}
    
\section{Introduction}
\label{sec:intro}

\begin{figure}[t!]
\centering
{
\includegraphics[width=\linewidth]{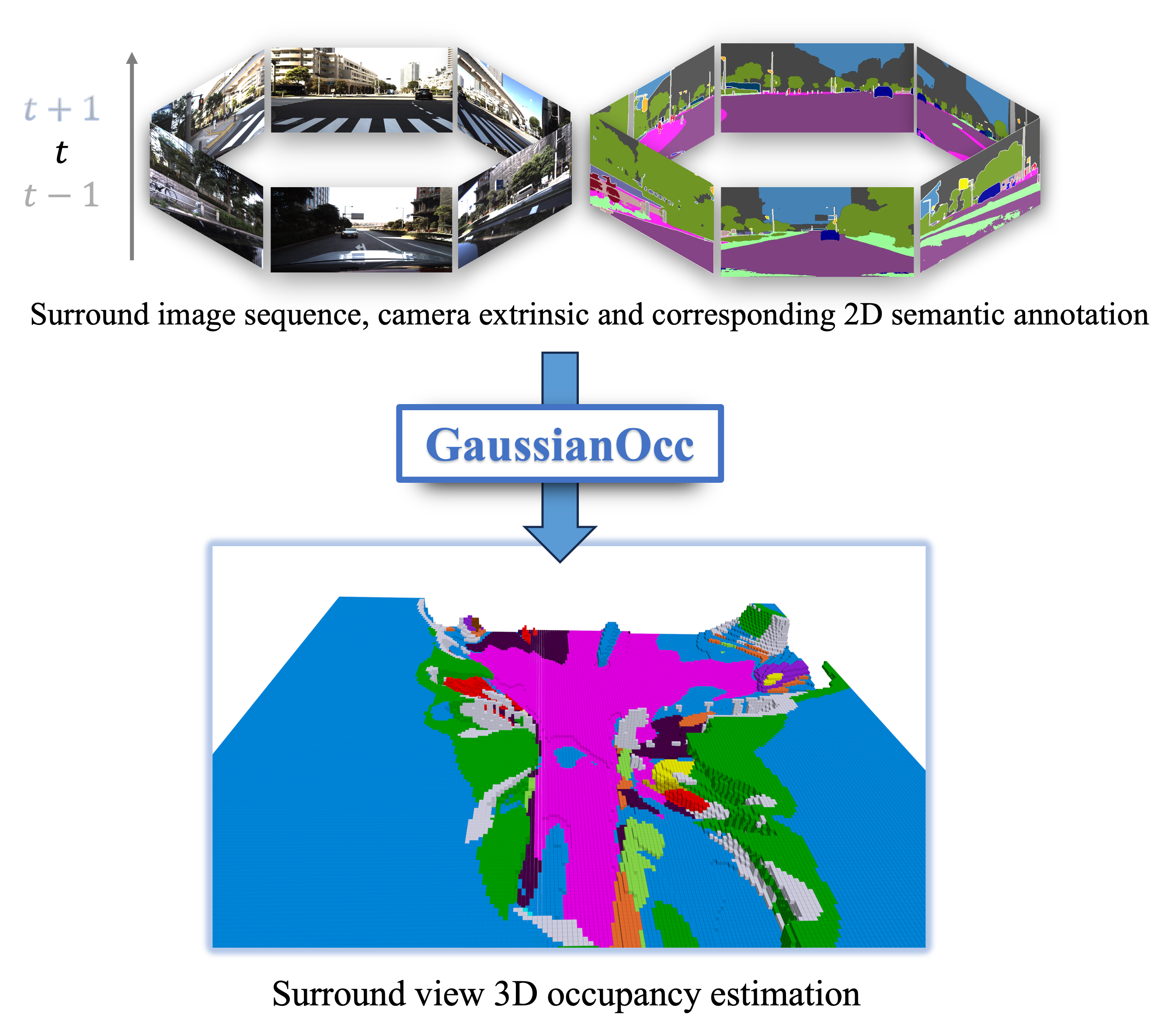}
}
 % \vspace{-25pt}
\caption{\textbf{Problem setting of GaussianOcc}. Given a surround image sequence, the spatial camera extrinsic and its corresponding 2D semantic annotation, GaussianOcc is able to perform 3D occupancy estimation without the need for ground truth occupancy label and ground truth 6D ego pose for training.}
\label{fig:task}
\end{figure}

% Vision-centric and grid-centric perception approaches have gained significant attention in the autonomous driving industry and academia \cite{ma2022vision, shi2023grid}. Among these, surround view 3D occupancy estimation \cite{sparseocc, surroundocc, flashocc,occsurvey} has emerged as a core perception task and a promising alternative to bird's-eye view (BEV) methods \cite{fang2023tbp, bevdepth, bevformer}.
% To facilitate 3D occupancy estimation, several benchmarks have been developed for supervised training \cite{surroundocc, openoccupancy, occ3d, tong2023scene}, though these require substantial effort in 3D annotation. To reduce the burden of annotation, self-supervised \cite{simpleocc, occnerf, selfocc, scenerf, hayler2023s4c} and weakly-supervised \cite{renderocc} learning approaches based on volume rendering have been proposed \cite{nerf, wimbauer2023behind, v4d}. Volume rendering allows 3D representation learning using 2D supervision signals, such as 2D semantic maps and depth maps, thereby eliminating the need for extensive 3D annotation.

Surround view 3D occupancy estimation \cite{sparseocc, surroundocc, flashocc,occsurvey, ma2022vision, shi2023grid, zhang2025visionpad} has emerged as a core perception task and a promising alternative to bird's-eye view (BEV) methods \cite{fang2023tbp, bevdepth, bevformer}.
To facilitate 3D occupancy estimation, several benchmarks have been developed for supervised training \cite{surroundocc, openoccupancy, occ3d, tong2023scene}, though these require substantial effort in 3D annotation. To reduce the burden of 3D annotation, self-supervised \cite{simpleocc, occnerf, selfocc, scenerf, hayler2023s4c} and weakly-supervised \cite{renderocc} learning approaches based on volume rendering have been proposed \cite{nerf, wimbauer2023behind, v4d}. Volume rendering allows 3D representation learning using 2D supervision signals, such as 2D semantic maps and depth maps, thereby eliminating the need for extensive 3D annotation.

Existing methods \cite{occnerf, selfocc} achieve self-supervised learning through volume rendering, where the 2D semantic map supervision is derived from open-vocabulary semantic segmentation \cite{zhang2023simple}, and the depth map supervision is obtained from self-supervised depth estimation \cite{simpleocc, monodepth2}. However, these approaches face two significant limitations. First, volume rendering is performed at real-world scale, which requires the ground truth 6D ego pose to calculate the multi-view photometric loss across sequential images. Second, the inefficiency in volume rendering poses a challenge, the same as in novel view synthesis tasks \cite{3dgs, 3dgssurvey, he2024neural}, due to the dense sampling operation required. These limitations impede the development of a more general and efficient paradigm for self-supervised 3D occupancy estimation.

To address the aforementioned limitations, we propose a fully self-supervised and efficient approach to 3D occupancy estimation with dedicated designs based on Gaussian splatting \cite{3dgs, 3dgssurvey}. Specifically, we introduce the use of Gaussian splatting to perform cross-view splatting, where the rendered image constructs a cross-view loss that provides scale information during joint training with the 6D pose network. This eliminates the need for ground truth 6D ego pose during training.
To improve rendering efficiency, we move away from the dense sampling required in traditional volume rendering. Instead, we propose performing Gaussian splatting directly from the 3D voxel space. In this approach, each vertex in the voxel grid is treated as a 3D Gaussian, and we optimize the attributes of these Gaussians—such as semantic and opacity—directly within the voxel space. Through the above innovative approach, our proposed method makes progress toward fully self-supervised and efficient 3D occupancy estimation, as outlined in Figure \ref{fig:task}.

In summary, our core contributions are as follows:
\begin{itemize}
    \setlength{\itemsep}{0pt}
    \setlength{\parsep}{0pt}
    \setlength{\parskip}{0pt}
    \setlength{\topsep}{0pt}
    \setlength{\partopsep}{0pt}
    \item We introduce the first fully self-supervised method for efficient surrounding-view 3D occupancy estimation, featuring the exploration of Gaussian splatting.
    \item We propose Gaussian splatting for cross-view projection module, which can provide scale information to get rid of the need of ground truth 6D ego pose during the training. 
    \item We propose Gaussian splatting from voxel space module, achieving competitive performance with 2.7 times faster training and 5 times faster rendering compared to the previous works with volume rendering. 

\end{itemize}

\section{Related work}
\label{sec:related_work}

%-------------------------------------------------------------------------
\subsection{Surround view depth estimation}
The surround view setting offers an ego-centric 360-degree perception solution \cite{ma2022vision, shi2023grid, nuscenes, ddad}. \cite{guo2023simple} introduces the surround view benchmark in the supervised setting, which learns the depth scale directly from the ground truth depth map. FSM \cite{fsm} is a pioneering work in scale-aware surround view depth estimation relying on stereo constraint \cite{yuan2024sd}. However, subsequent studies \cite{surrounddepth, kim2022self} have found that reproducing the performance of \cite{fsm} is challenging. Surrounddepth \cite{surrounddepth} improves the scale supervision signal using sparse point clouds from Structure-from-Motion (SFM). Building on the spatial and temporal constraints in FSM, \cite{kim2022self} introduces a volume feature fusion module to enhance performance. For better performance, \cite{r3d3} proposes the temporal offline refinement strategy based on the multiple cameras and monocular depth refinement.
These works \cite{fsm, surrounddepth, kim2022self, r3d3} all use traditional projection and index interpolation for cross-view synthesis, computing the loss between the synthesized view and target images. Compared to the traditional projection, our approach employs Gaussian splatting with a dedicated design for cross-view constraint, achieving better performance.

\subsection{Surround view 3D occupancy estimation}

Surround view 3D occupancy estimation has gained significant attention in recent years, with several benchmarks based on the nuScenes dataset \cite{surroundocc, occ3d, openoccupancy}. In addition to the advanced architectures being proposed \cite{sparseocc, occformer, flashocc, fbocc}, another research trend involves utilizing volume rendering for 3D occupancy learning with 2D supervision \cite{simpleocc, renderocc, occnerf, selfocc}. SimpleOcc \cite{simpleocc} pioneered the use of volume rendering for 3D occupancy estimation, exploring both supervised and self-supervised learning. RenderOcc \cite{renderocc} extends semantic information for rendering. OccNeRF \cite{occnerf} and SelfOcc \cite{selfocc} share a similar approach by using 2D open-vocabulary semantic models to generate semantic maps for supervision. However, since volume rendering processes are conducted at real-world scale, these self-supervised methods \cite{simpleocc, occnerf, selfocc} require ground truth 6D poses from sensors to provide the real-world scale for training. Differently, we are exploring a solution that utilizes the overlap region in adjacent cameras to learn the real-world scale, eliminating the need for ground truth 6D poses.

\begin{figure*}
    \centering
    \includegraphics[width=1.0 \textwidth]{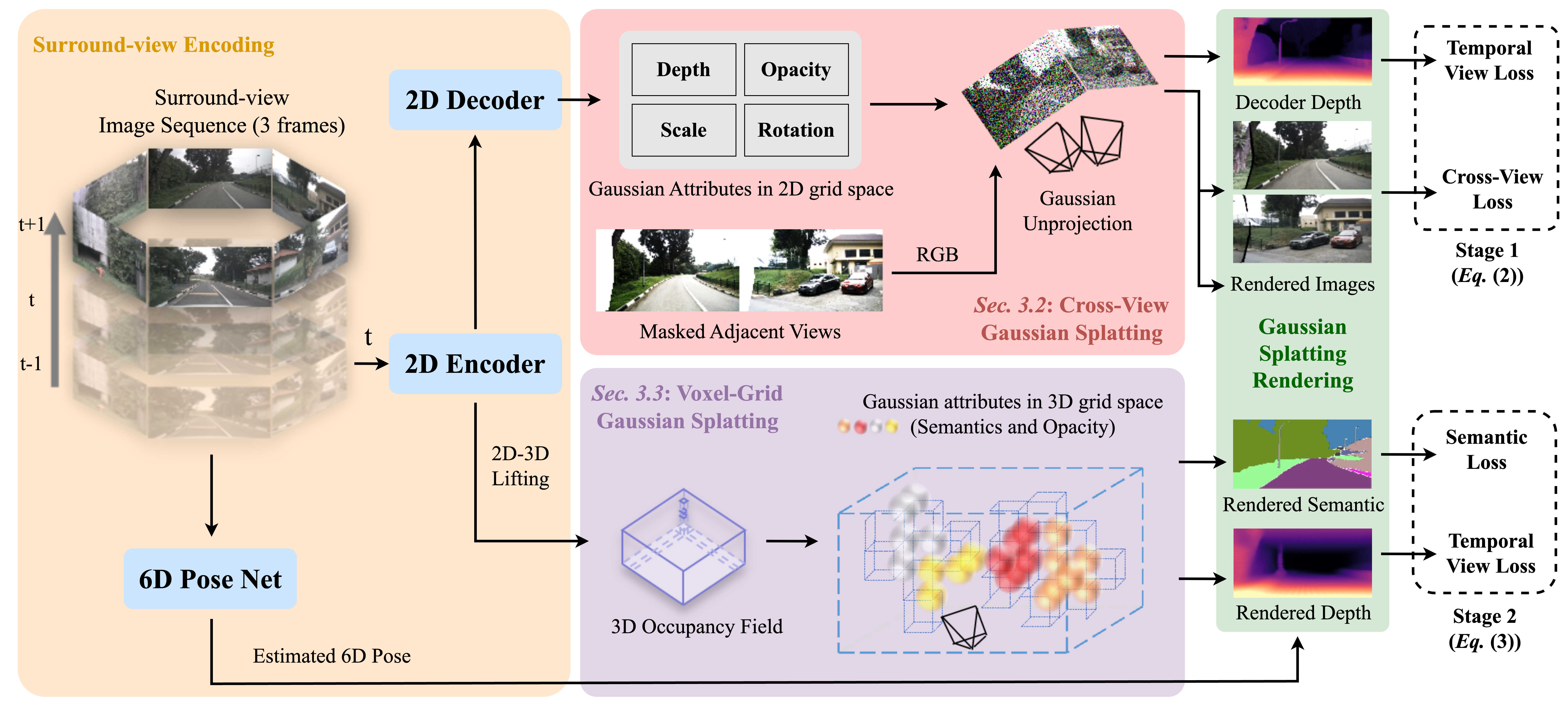}
     \vspace{-10pt}
    \caption{\textbf{GaussianOcc is a two-stage method.} In Stage 1, we train a scale-aware 6D pose network, using a U-Net architecture to predict Gaussian attributes in the 2D image grid space for cross-view Gaussian splatting. This approach provides scale information in the joint training with the 6D pose net. Based on the 6D pose from Stage 1, we perform self-supervised 3D occupancy estimation in Stage 2, where we lift the 2D features to a 3D voxel space and propose voxel grid Gaussian splatting for fast rendering. \textit{Note that, for clarity, we omit the line from the 6D pose network to the loss in Stage 1, and the 2D encoder is independent for each stage (not shared).}}
    \label{fig:Overview}
\end{figure*}

\subsection{3D Gaussian splatting}

3D Gaussian splatting has become a popular method for modeling 3D and 4D scenes using well-posed images \cite{3dgs, katsumata2023efficient, duan20244d, wu20244d}, which has the property of fast rendering compared to the volume rendering in neural radiance field \cite{nerf, he2024neural, v4d}. In driving scenes, a line of research has focused on scene-specific reconstruction \cite{s3Gaussian, drivinggaussian, yan2024street}. Our work, however, investigates the function of Gaussian splatting in a generalized setting, where existing methods generally construct 3D Gaussians from the unprojection of learned 2D Gaussian attributes \cite{pixelsplat, zheng2024gps, liu2024fast, szymanowicz2024splatter}. We also employ this unprojection approach, but uniquely, our approach constructs cross-view information from adjacent views to learn scale information through Gaussian splatting projection. In addition, we investigate the Gaussian splatting on the voxel space for faster rendering compared with the previous volume rendering-based works \cite{simpleocc, occnerf, selfocc}.

Note that, recent works, GaussianFormer \cite{gaussianformer}, GaussTR \cite{GaussTR} and GaussianBeV \cite{gaussianbev}, are related to ours in their focus on 3D occupancy estimation and BEV prediction. However, our exploration diverges by focusing on two new properties that Gaussian splatting can contribute to occupancy estimation: scale-aware training and faster rendering.

\section{Method}
\label{sec:formatting}

%-------------------------------------------------------------------------
\subsection{Preliminaries}

3D Gaussian splatting \cite{3dgs} is for modeling static 3D scenes using point primitives, where each primitive is characterized with the following attributes: (1) a 3D position $\mathcal{X} \in \mathbb{R}^3$, (2) a color defined by SH coefficients $c \in \mathbb{R}^k$ (where $k$ denotes the dimensionality of the SH basis), (3) a rotation represented by a quaternion $r \in \mathbb{R}^4$, (4) a scaling factor $s \in \mathbb{R}^3_+$, and (5) an opacity $\alpha \in [0, 1]$. The original Gaussian splatting is for scene-specific, fast 3D novel view synthesis, where the attributes of Gaussian points are optimized by the multi-view constraint. Differently, we study fully self-supervised and efficient 3D occupancy estimation by exploring the Gaussian attributes that are well-aligned in both 2D and 3D grids. Our design allows us to benefit from the Gaussian splatting rendering for the scale-aware training by cross-view constraint and faster rendering on voxel grids, as illustrated in Figure \ref{fig:Overview}. 

% This approach models the 3D scene in the 2D image plane as a depth map and in 3D grid space as a voxel (occupancy) format.

% Each point is represented as a scaled Gaussian distribution characterized by a 3D covariance matrix $\Sigma$ and a mean $\mu$. The Gaussian distribution for a point $\mathcal{X}$ is given by:

% \begin{equation}
%     G(\mathcal{X}) = e^{-\frac{1}{2} (\mathcal{X} - \mu)^T \Sigma^{-1} (\mathcal{X} - \mu)}
% \end{equation}

% To enable efficient optimization via gradient descent, the covariance matrix $\Sigma$ is decomposed into a scaling matrix $\mathbf{S}$ and a rotation matrix $\mathbf{R}$ as follows:

% \begin{equation}
%     \Sigma = \mathbf{R S S}^T \mathbf{R}^T
% \end{equation}

% The projection of Gaussians from 3D space onto a 2D image plane involves a view transformation $\mathbf{W}$ and the Jacobian of the affine approximation of the projective transformation $\mathbf{J}$. The 2D covariance matrix $\Sigma'$ is computed as:

% \begin{equation}
%     \Sigma' = \mathbf{J W \Sigma W}^T \mathbf{J}^T
% \end{equation}

% Subsequently, an alpha-blend rendering technique, similar to that used in NeRF \cite{nerf}, is applied. This is formulated as:

% \begin{equation}
%     \mathbf{C}_{\text{color}} = \sum_{i \in N} c_i \alpha_i \prod_{j=1}^{i-1} (1 - \alpha_i)
% \end{equation}

% Here, $c_i$ represents the color of each point, and the density $\alpha_i$ is determined by the product of a 2D Gaussian with covariance $\Sigma'$ and a learned per-point opacity. The color is defined using spherical harmonics (SH) coefficients as described in \cite{plenoxels, 3dgs}.

\subsection{Scale-aware training by Gaussian Splatting}

\noindent \textbf{Scale from spatial camera rig:} Similar to the previous work \cite{fsm, surrounddepth, kim2022self}, the scale information is from the surround camera rig. Specifically, the real-world scale can be obtained by leveraging camera extrinsic matrices, which is to use spatial photometric loss in the overlap region between two adjacent views, \ie, warping ${I}^{i}_t$ to ${I}^{j}_t$:
\begin{equation}
    p^{i \rightarrow j}_t = K^{j} (T^{j})^{-1} T^{i} D_t^{i} (K^{i})^{-1} p^{i}_t,
\end{equation}
where $K^{i}, T^{i}$ are the intrinsic and extrinsic matrices of $i$-th camera, $D_t^{i}$ is the predicted depth map of $i$-th camera, $p_t$ is the corresponding pixel during the warping. The warping operation is achieved by direct bilinear interpolation with the corresponding $p^{i \rightarrow j}_t $. However, as pointed out by \cite{surrounddepth}, the mapping in such a small overlap region, $p^{i \rightarrow j}_t$, can easily go into sub-optimal depth result, where we verify this in the experiment section in Figure \ref{fig:cross_view}. Therefore, apart from the spatial loss, \cite{surrounddepth} proposes to facilitate the Structure-from-Motion (SFM) to extract sparse depth information for direct depth supervision to provide a stronger supervision signal, but it is time-consuming and not straightforward. Different from \cite{surrounddepth}, \cite{kim2022self} enhances the depth estimation performance with spatio-temporal context that does not need the sparse depth from SFM but the performance is still limited. Inspired by the explicit sparse depth supervision in \cite{surrounddepth}, we ask whether we can enforce the cross-view constraint on adjacent views more explicitly. The answer is yes. We find that the nature of Gaussian splatting is scale-aware projection that could serve for the cross-view stereo constraint. We propose Gaussian splatting for projection in stage 1 for better scale-aware training as follows. 

\noindent \textbf{Gaussian Splatting for Projection (GSP):} As illustrated in Figure \ref{fig:Overview}, we adapt a depth network \cite{newcrfs} to predict the Gaussian attributes in 2D grid space, where, apart from the original depth map, we also predict the scale map and rotation map. For each adjacent view, we first calculate the mask in the overlap region, then mask out one side of these overlap regions. Due to the presence of the other side's overlap region, the unprojected 3D scene remains complete if the depth map is predicted well. This mask-out step is critical for providing scale training, as indicated in the experiment section (Table \ref{tab:scale traning}). We then perform splatting rendering on the adjacent views to obtain the rendered image. If the depth map is accurately learned, the rendered image should resemble the original images, providing the necessary scale information for the joint training with 6D pose net. 

\begin{figure}[t!]
\centering
{
\includegraphics[width=\linewidth]{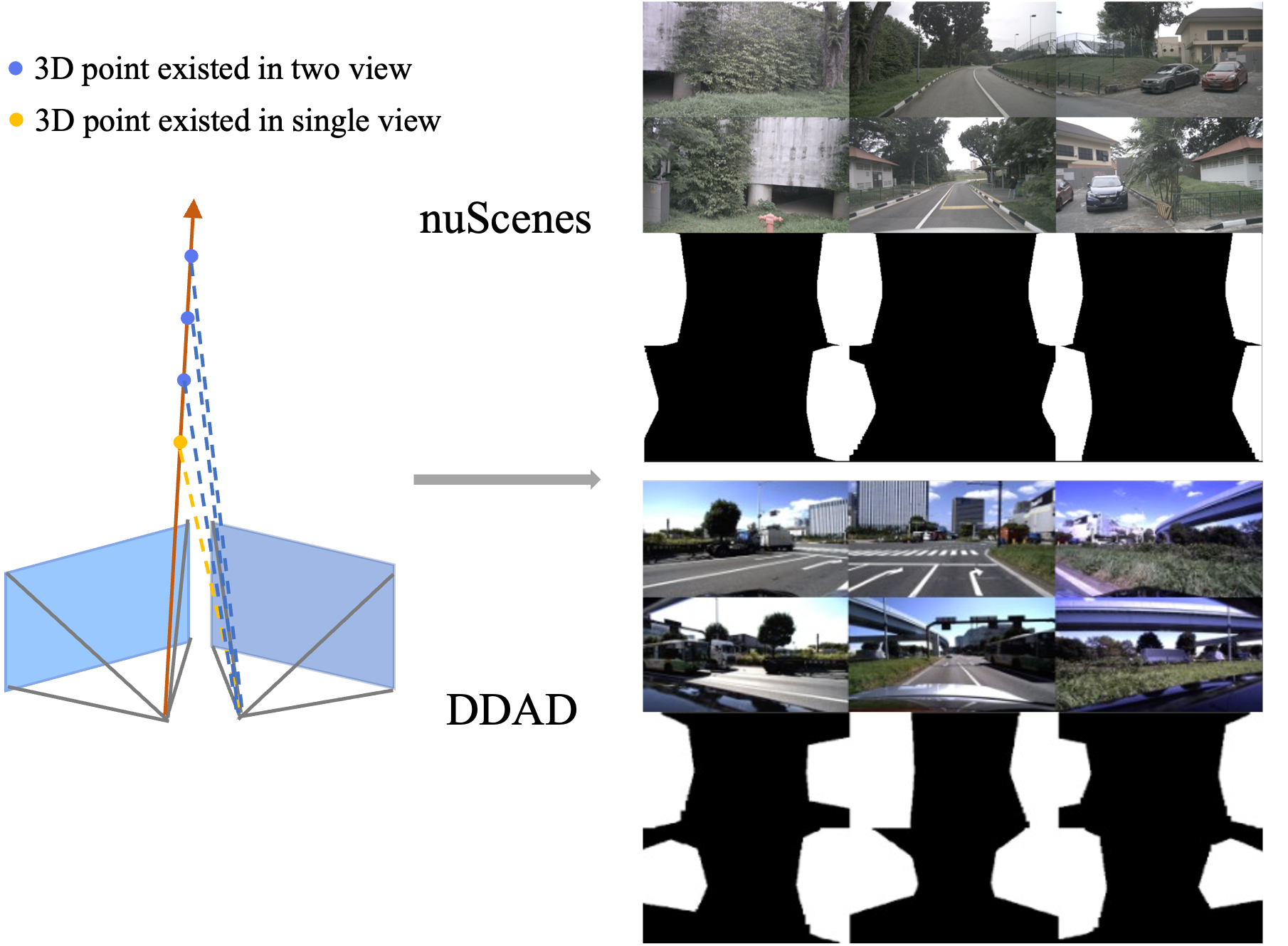}
}
 \vspace{-15pt}
\caption{\textbf{Overlap mask in nuScenes \cite{nuscenes} and DDAD \cite{ddad}}. }
\label{fig:mask}
 \vspace{-5pt}
\end{figure}

\noindent \textbf{Overlap Mask:} The process of acquiring the overlap mask is illustrated in Figure \ref{fig:mask}. We densely sample points along the ray in one view, and a pixel is considered part of the overlap region if more than one sampled 3D point falls within the adjacent view. The overlap mask is only determined by the camera's extrinsic and defined max depth (e.g., 80m in nuScenes). Note that in the DDAD dataset \cite{ddad}, we exclude regions with self-occlusion (such as parts of the vehicle body). Besides, we apply the erosion operation from OpenCV \cite{opencv} to the mask for purification.

\subsection{Fast rendering by Gaussian Splatting}

\noindent \textbf{Inefficient performance in volume rendering:} For 2D supervision (semantic and depth maps), previous methods \cite{simpleocc, occnerf, selfocc, renderocc} employed volume rendering based on dense sampling. Although the final 3D voxel representation for modeling the 3D scene is much quicker than the original implicit representation \cite{nerf}, it remains time-consuming, particularly when incorporating semantic map rendering. For example, in OccNeRF \cite{occnerf}, the number of sampled points at a resolution of $180 \times 320$ is 108,735,066. However, the target optimized points correspond to the vertices in the 3D voxel grid, totaling $300 \times 300 \times 24 = 2,160,000$. This redundancy in densely sampled points helps optimization with volume rendering but is highly inefficient.

\noindent \textbf{Gaussian Splatting from Voxel (GSV):} As analyzed above, the target optimized points are the vertices in the 3D voxel grid, prompting us to consider directly optimizing these vertices. Interestingly, we find that it is suitable to use the Gaussian splatting to replace the volume rendering if we regard the vertices of each the voxel grid as the position of the 3D Gaussian. Then, we can optimize the attributes of the 3D Gaussian, such as semantic and opacity information. For example, in the empty space of voxel, even though we have the vertices at that region during the splatting rendering, after the optimization, the network would predict the opacity as zero at these vertices then these vertices would not contribute any geometry or semantic information during the rendering. Since all vertices are arranged in 3D voxel space with real-world 3D position $\mathcal{X}$, we can use fixed scale $s$ and fixed rotation $r$ for each vertex for simplification. This allows us to model the 3D scene by optimizing the rest Gaussian attributes (semantic and opacity).

\subsection{Loss function}

We formulate the loss function of each stage as follows: 
\begin{equation}
\label{total loss}
\mathcal{L}_{stage 1}=\mathcal{L}_{temporal} + \mathcal{L}_{cross},
\end{equation}
\begin{equation}
\mathcal{L}_{stage 2}=\mathcal{L}_{temporal} + \lambda \mathcal{L}_{semantic},
\end{equation}
\begin{equation}
(\mathcal{L}_{temporal}, \mathcal{L}_{cross})= \frac{1-\operatorname{SSIM}\left(I_t, \hat{I}_t\right)}{2}+ \beta \left\|I_t-\hat{I}_t\right\|,
\end{equation}
where \(I_t\) and \(\hat{I}_t\) refer to the target image and the corresponding synthesized image, respectively. Note that \(\hat{I}_t\) in the temporal-view photometric loss \(\mathcal{L}_{temporal}\) is generated by projecting pixels from the source image using the coordinate index.  In contrast, \(\hat{I}_t\) in the cross-view photometric loss \(\mathcal{L}_{cross}\) is derived from our proposed cross-view Gaussian splatting method. \(\mathcal{L}_{semantic}\) is 2D semantic loss with balanced weight $\lambda = 0.02$. 
For \(\mathcal{L}_{cross}\) and \(\mathcal{L}_{temporal}\), we set \(\beta\) is set to 0.15 for weight balance the same as~\cite{occnerf}. 

% For \(\mathcal{L}_{temporal}\), we recommend these works \cite{monodepth2, simpleocc, occnerf} for further details. For \(\mathcal{L}_{semantic}\), we recommend \cite{occnerf} for a better understanding. 

\definecolor{nbarrier}{RGB}{255, 120, 50}
\definecolor{nbicycle}{RGB}{255, 192, 203}
\definecolor{nbus}{RGB}{255, 255, 0}
\definecolor{ncar}{RGB}{0, 150, 245}
\definecolor{nconstruct}{RGB}{0, 255, 255}
\definecolor{nmotor}{RGB}{200, 180, 0}
\definecolor{npedestrian}{RGB}{255, 0, 0}
\definecolor{ntraffic}{RGB}{255, 240, 150}
\definecolor{ntrailer}{RGB}{135, 60, 0}
\definecolor{ntruck}{RGB}{160, 32, 240}
\definecolor{ndriveable}{RGB}{255, 0, 255}
\definecolor{nother}{RGB}{139, 137, 137}
\definecolor{nsidewalk}{RGB}{75, 0, 75}
\definecolor{nterrain}{RGB}{150, 240, 80}
\definecolor{nmanmade}{RGB}{230, 230, 250}
\definecolor{nvegetation}{RGB}{0, 175, 0}
\definecolor{nothers}{RGB}{0, 0, 0}

\definecolor{col1}{RGB}{232, 161, 148}
\definecolor{col2}{RGB}{148, 187, 232}

\section{Experiment}

\subsection{Tasks, datasets, and metric}

\noindent \textbf{nuScenes \cite{nuscenes}:} For 3D occupancy estimation, we utilize annotations from Occ3D~\cite{occ3d}. For fair comparison, we use the 2D pseudo semantic map provided by OccNeRF \cite{occnerf} for training. We measure 3D occupancy estimation performance using the mean Intersection over Union (mIoU) metric. For depth estimation, we set the perception range in [-80m, -80m, -1m, 80m, 80m, 6m], while we clamp the ground truth to a range of 0.1m to 80m for evaluation, consistent with OccNeRF and SurroundDepth \cite{surrounddepth}. We evaluate depth maps using error metrics (Abs Rel, Sq Rel, RMSE, RMSE log) and threshold accuracy metrics ($\delta$). 
% We do not use the multi-frame rendering used in \cite{occnerf} as we do not observe the improvement.

\noindent \textbf{DDAD \cite{ddad}:} Though we do not have the 3D occupancy labels on DDAD dataset, we can present qualitative results thanks to our fully self-supervised 3D occupancy estimation setting. We obtain the 2D pseudo semantic labels for training following OccNeRF pipeline in nuScenes dataset. For depth estimation, 
we clamp the depth range within 0.1m and 200m for evaluation, consistent with SurroundDepth \cite{surrounddepth}. 

% we also set the perception range in [-80m, -80m, -1m, 80m, 80m, 6m], but clamp the ground truth to a range of 0.1m to 200m for evaluation, consistent with SurroundDepth \cite{surrounddepth}. 

% \noindent \textbf{DDAD \cite{ddad}:} Though we do not have the 3D occupancy labels on DDAD dataset, we can present qualitative results thanks to our fully self-supervised 3D occupancy estimation setting. We obtain the 2D pseudo semantic labels for training following the OccNeRF pipeline in nuScenes dataset. Besides, we also provide the depth estimation result in supplementary material.

% For depth estimation, we also set the perception range in [-80m, -80m, -1m, 80m, 80m, 6m], but clamp the ground truth to a range of 0.1m to 200m for evaluation, consistent with SurroundDepth \cite{surrounddepth}. 

\begin{table}[t]
	% \footnotesize
 % 	\setlength{\tabcolsep}{0.0025\linewidth}
	% \newcommand{\classfreq}[1]{{~\tiny(\nuscenesfreq{#1}\%)}}  %
    \begin{center}
	\resizebox{0.48\textwidth}{!}{
	\begin{tabular}{l|c c|c c}
		\toprule
		Method & {GT Occ.} & {GT Pose} & mIoU* & mIoU
        
		\\
		\midrule
    MonoScene~\cite{monoscene} & \checkmark & $\times$ & 6.33 & 6.06    \\
    BEVDet ~\cite{bevdet} & \checkmark & $\times$ & 20.03 & 19.38  \\
    BEVFormer~\cite{bevformer} & \checkmark & $\times$ & 24.64 & 23.67  \\
    OccFormer~\cite{occformer}& \checkmark & $\times$ & 22.39& 21.93  \\
    TPVFormer~\cite{tpvformer} &\checkmark & $\times$ & 28.69 & 27.83  \\
    CTF-Occ~\cite{occ3d} & \checkmark & $\times$ & \bf{29.54} & \bf{28.53} \\
     \midrule
    RenderOcc~\cite{renderocc} & $\times$ & $\times$ &  24.53 &23.93 \\ 
        \midrule
        SimpleOcc~\cite{simpleocc} & $\times$ & \checkmark  & 7.99 & 7.05  \\
         SelfOcc~\cite{selfocc} &  $\times$ & \checkmark & 10.54  & 9.30 
         \\
         OccNeRF~\cite{occnerf} &  $\times$ & \checkmark & 10.81  & 9.54  \\

          GaussianOcc &  $\times$ & $\times$ & \bf{11.26}  & \bf{9.94}  \\
          
	\bottomrule
	\end{tabular}}
    \end{center}
     \vspace{-12pt}
    \caption{\textbf{3D occupancy comparison on the Occ3D dataset with mIoU metric.} Since `other' and `other flat' classes are the invalid prompts for open-vocabulary models, we also calculate `mIoU*' as the result ignoring the classes that do not consider these two classes during evaluation, while `mIoU' is the original result. GT Occ. means using the ground truth occupancy label for supervision. GT Pose is the ground truth 6D ego pose from the sensor for self-supervised geometry learning.}
	\label{tab:occ}

\end{table}

\begin{table*}[tb]
	\centering
	\resizebox{0.95\textwidth}{!}{
	\begin{tabular}{l|c|c|c|c|c|c|c|c|c}
			\hline
			Method & \textit{GT pose} & \textit{Occ.} &\cellcolor{col1}Abs Rel & \cellcolor{col1}Sq Rel & \cellcolor{col1}RMSE  & \cellcolor{col1}RMSE log & \cellcolor{col2}$\delta < 1.25 $ & \cellcolor{col2}$\delta < 1.25^{2}$ & \cellcolor{col2}$\delta < 1.25^{3}$\\
           \hline
        \multicolumn{10}{c}{nuScenes~\cite{nuscenes}}
\\
 \hline
			FSM~\cite{fsm} & $\times$ & $\times$ &  0.297  &   -  &   -  &   -  &   -  &   -  &   -
\\

			FSM*~\cite{fsm}& $\times$ & $\times$ &   0.319 &7.534 &7.860 &0.362 &0.716 &0.874 &0.931

 \\
			SurroundDepth~\cite{surrounddepth} &  $\times$ & $\times$ & 0.280 & \bf{4.401} &7.467 &0.364 &0.661 &0.844& 0.917
\\            
            SA-FSM~\cite{SA-FSM} & $\times$ & $\times$ & 0.272 & 4.706 & 7.391 & 0.355 & 0.689 & 0.868 & 0.929

\\
            VFF~\cite{kim2022self} & $\times$ & $\times$ &  0.289 &5.718 &7.551 &0.348 &0.709 &0.876 &0.932

\\
            R3D3~\cite{r3d3} & $\times$ & $\times$ &  \bf{0.253} &4.759 & \bf{7.150} &- &0.729 &- &-
            
\\

		  GaussianOcc $\ddagger$  & $\times$ & $\times$ &  0.258  &   5.733  &   7.222  &   \bf{0.343}  &   \bf{0.753}  &   \bf{0.888}  &   \bf{0.934}  

\\
\hline
            SimpleOcc~\cite{simpleocc} & \checkmark & \checkmark &  0.224 &3.383 & 7.165 & 0.333 &  0.753 &0.877 &0.930
\\
			  OccNeRF~\cite{occnerf} & \checkmark & \checkmark & 0.202  &   2.883  &  \bf{6.697}  &   0.319  &   \bf{0.768}  &   \bf{0.882}  &  0.931
\\
     		OccNeRF~~\cite{occnerf} $\dagger$ & \checkmark & \checkmark &  0.456  &  12.682  &   9.194  &   0.399  &   0.704  &   0.833  &   0.890 

\\			  SelfOcc~\cite{selfocc} & \checkmark & \checkmark & 0.215 & 2.743 & 6.706 & 0.316 & 0.753 & 0.875 & \bf{0.932}

\\
		   GaussianOcc & $\times$ & \checkmark &  0.211  &   3.115  &   7.131  &   0.326  &   0.762  &   0.878  &   0.931
\\
		   GaussianOcc $\dagger$ & $\times$ & \checkmark &  \bf{0.197}  &   \bf{1.846}  &   6.733  &   \bf{0.312}  &   0.746  &   0.873  &   0.931  
\\
\hline

        \multicolumn{10}{c}{DDAD~\cite{ddad}}
\\
 \hline

        FSM* ~\cite{fsm} & $\times$ & $\times$ &  0.228 & 4.409 & 13.433 & 0.342 & 0.687 & 0.870 & 0.932
\\

        VFF~\cite{kim2022self} & $\times$ & $\times$ &  0.218 & 3.660 & 13.327 & 0.339 & 0.674 & 0.862 & 0.932
\\

        SurroundDepth~\cite{surrounddepth} & $\times$ & $\times$ & 0.208  & 3.371  & 12.977  & 0.330  & 0.693  & 0.871  & 0.934
\\        
        SA-FSM~\cite{SA-FSM} & $\times$ & $\times$ & 0.187 & 3.093 & 12.578 & \bf{0.311} & 0.731 & \bf{0.891} & \bf{0.945}
\\
        R3D3~\cite{r3d3} & $\times$ & $\times$ &  \bf{0.162} & \bf{3.019} & \bf{11.408} &- & \bf{0.811} &- &-
\\
	GaussianOcc $\ddagger$  & $\times$ & $\times$ &  0.212  &   3.556  &  12.564  &   0.320  &   0.701  &   0.888  &   0.944

\\
\hline
	GaussianOcc & $\times$ & \checkmark &  0.228  &   3.854  &  14.326  &   0.357  &   0.660  &   0.853  &   0.922
\\
 \hline
	\end{tabular}}
  \vspace{-5pt}
	\caption{\textbf{Comparisons for self-supervised multi-camera depth estimation on the nuScenes ~\cite{nuscenes} and DDAD datasets ~\cite{ddad}}. The results are averaged over all views without median scaling at test time. `FSM*' is the reproduced result in~\cite{kim2022self}. GaussianOcc $\ddagger$ represents the depth estimation result from Stage 1. GaussianOcc $\dagger$ and OccNeRF $\dagger$ means the model trained with the semantic information. \textit{Occ.} represents the ability of the method to predict the 3D occupancy.}
	\label{tab:depth benckmark}
\end{table*}

\begin{figure*}
    \centering
    \includegraphics[width=0.90 \textwidth]{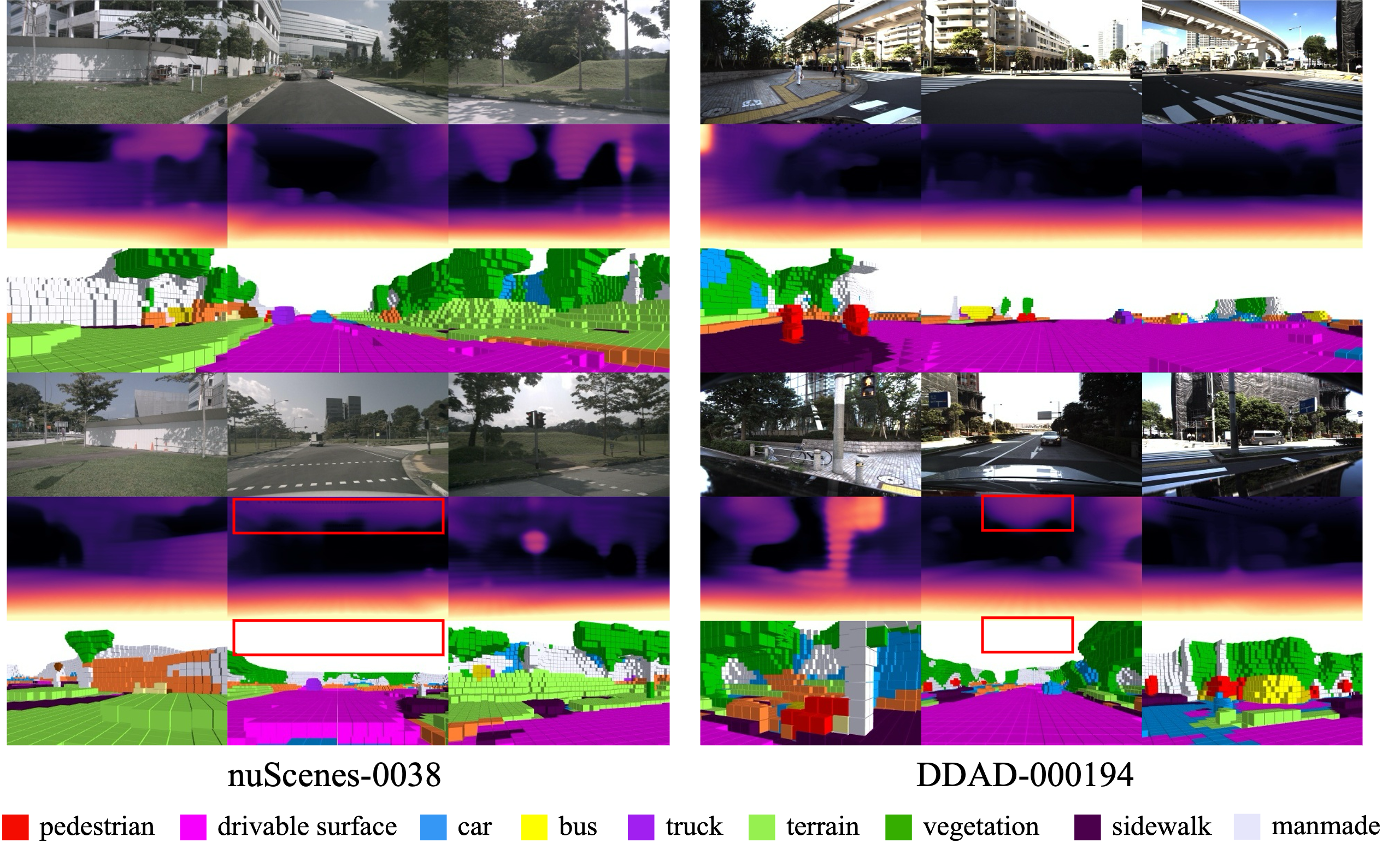}
     \vspace{-10pt}
    \caption{\textbf{Visualization of the render depth map and 3D occupancy prediction on the nuScenes and DDAD datasets.} }
    \label{fig:occresult}
\end{figure*}

\begin{table*}[t]
	\centering
	\resizebox{1.0\textwidth}{!}{
		\begin{tabular}{c|c|c|c|c|c|c|c|c|c|c|c|c}
			\hline
			\multicolumn{2}{c|}{Scale-aware training in \cite{surrounddepth, kim2022self}}  &  \multicolumn{4}{c|}{Scale-aware training by ours } &\cellcolor{col1} & \cellcolor{col1} & \cellcolor{col1}  & \cellcolor{col1}RMSE  & \cellcolor{col2} & \cellcolor{col2} & \cellcolor{col2}\\
			
			\textit{Loss in \cite{surrounddepth}} & \textit{Loss in \cite{kim2022self} } & \textit{GS loss} & \textit{Mask} & \textit{Erode} & \textit{Refine} & \multirow{-2}{*}{\cellcolor{col1}Abs Rel}&\multirow{-2}{*}{\cellcolor{col1}Sq Rel} & \multirow{-2}{*}{\cellcolor{col1}RMSE}  & \cellcolor{col1}log & \multirow{-2}{*}{\cellcolor{col2}$\delta < 1.25 $} & \multirow{-2}{*}{\cellcolor{col2}$\delta < 1.25^{2}$} & \multirow{-2}{*}{\cellcolor{col2}$\delta < 1.25^{3}$}\\
			\hline
                \checkmark* &  & && &  &  0.280 &  \bf{4.401} &  7.467 &  0.364 &  0.661 &  0.844 &  0.917 
			\\
               \checkmark &  & & &&  & 0.672  &  25.405  &  11.999  &   0.568  &   0.419  &   0.808  &   0.878
			\\
               & \checkmark* & & & & &  0.289 &5.718 &7.551 &0.348 &0.709 &0.876 &0.932
			\\
                &  \checkmark & & &  & & 0.285 & 6.046 & 7.514 & \bf{0.342} & 0.702 & 0.865 & 0.931
			\\
            \hline
   	  &  &  \checkmark & &  & &  0.798  &  11.571  &  15.251  &   1.472  &   0.006  &   0.015  &   0.028
			\\

   	   & & \checkmark & \checkmark &  & &  0.293  &   7.127  &   7.536  &   0.376  &   0.743  &   0.876  &   0.923
			\\
           & & \checkmark & \checkmark & \checkmark & & 0.281  &   6.986  &   7.347  &   0.354  &   \bf{0.766}  &   0.885  &   0.929
			\\
            & & \checkmark & \checkmark & \checkmark & \checkmark & \bf{0.258}  &   5.733  &   \bf{7.222}  &   0.343  &   0.753  &   \bf{0.888}  &   \bf{0.934}  

			\\
   
		\hline
   
	\end{tabular}}
	% \vspace{1pt}
	\caption{Ablation study for scale-aware depth estimation on the nuScenes dataset \cite{nuscenes}.  \checkmark* means the result from the original paper and \checkmark means the result using New-CRFs \cite{newcrfs} as the depth network the same as ours. \textit{GS loss} means using the spatial context constraint by our proposed Gaussian splatting for projection. \textit{Mask} represents using the mask-out strategy before the unprojection. \textit{Erode} means the erode process to the binary overlap mask and \textit{Refine} is the refinement of depth estimation network with 2 epochs by fixing the 6D pose net.}
	% \vspace{-5mm}
	\label{tab:scale traning}
\end{table*}

\subsection{Implementation details}

\noindent \textbf{Network details:}  For U-Net architecture, we adapt New-CRFs \cite{newcrfs} to predict the Gaussian attributes, which is based on the Swin Transformer \cite{swin}. The 6D pose net is the same as that used in SurroundDepth \cite{surrounddepth}. For the 2D-to-3D lifting, we follow the approach used in SimpleOcc \cite{simpleocc}. In the depth estimation benchmark, we use the network proposed by SimpleOcc, where the final output size is 256×256×16. In our Gaussian splatting setting, we further upsample the final output to 
512×512×32 for improved performance since we observe that a finer voxel grid leads to a finer rendered depth map, which requires ignorable computational cost. For occupancy estimation, we use the same network as OccNeRF \cite{occnerf} to ensure a fair comparison.

\noindent \textbf{Training details:} We propose a two-stage training for fully self-supervised 3D occupancy estimation as indicated in Figure \ref{fig:Overview}. In stage 1, we jointly train the depth estimation network and the 6D pose net, where we train the models for 8 epochs on the nuScenes and 12 epochs on the DDAD. In stage 2, we train the 3D occupancy network in a self-supervised manner with the 6D pose predicted from stage 1 rather than the ground truth pose used in OccNeRF \cite{occnerf}. We train the models for 12 epochs on both the nuScenes and DDAD. The optimizer and learning rate adjustment strategy follow those used in SimpleOcc \cite{simpleocc} and OccNeRF \cite{occnerf}.

\subsection{Main results}

% \noindent \textbf{3D occupancy estimation in mIoU metric:} As shown in Table \ref{tab:occ}, the proposed GaussianOcc achieves the best performance compared with other self-supervised methods, without the need for ground truth occupancy labels and ground truth 6D poses for training. 
% It's worth noting that SelfOcc \cite{selfocc} differs slightly from methods like SimpleOcc, OccNeRF, and our approach. 
% Unlike these methods, SelfOcc does not predict semantic information directly in the 3D occupancy space. Instead, it attaches 2D semantic information to the 3D occupied voxels, where the 2D semantic data is sourced from a third-party open-vocabulary model \cite{zhang2023simple}. 

% \noindent \textbf{3D occupancy estimation in mIoU and RayIoU metric:} As shown in Table \ref{tab:occ}, the proposed GaussianOcc achieves the best performance compared with other self-supervised methods, without the need for ground truth occupancy labels and ground truth 6D poses for training.  We also compare our method using a newly proposed metric, RayIoU, introduced by the recent work \cite{sparseocc}. The RayIoU is a ray-based evaluation metric to solve the inconsistency penalty along the depth axis raised in traditional voxel-level mIoU criteria. As shown in Table \ref{table:rayiou}, our approach also outperforms OccNeRF \cite{occnerf}. It's important to note that the FPS is calculated excluding rendering time. Since GaussianOcc and OccNeRF utilize the same network architecture, they share the same inference time when the rendering process is not taken into account.

\noindent \textbf{3D occupancy estimation in nuScenes:} In Table \ref{tab:occ}, the proposed GaussianOcc achieves the best performance compared to other self-supervised methods. In particular, we do not require ground truth occupancy labels and ground truth 6D ego pose for training, thanks to the predicted 6D pose from the stage 1. Note that RenderOcc \cite{renderocc} does not require the 3D occupancy label, but it is not a self-supervised method since it uses the ground truth depth map and semantic map for the 2D supervision; it could be regarded as a weakly-supervised method. In addition, we also achieve the best result on the RayIoU metric \cite{sparseocc}, with further details provided in the supplementary material.

% We also compare our method using a newly proposed metric, RayIoU, introduced by the recent work \cite{sparseocc}. The RayIoU is a ray-based evaluation metric to solve the inconsistency penalty along the depth axis raised in traditional voxel-level mIoU criteria. As shown in Table \ref{table:rayiou}, our approach also outperforms OccNeRF \cite{occnerf}. It's important to note that the FPS is calculated excluding rendering time. Since GaussianOcc and OccNeRF utilize the same network architecture, they share the same inference time when the rendering process is not taken into account.

\noindent \textbf{3D occupancy estimation in DDAD:} To the best of our knowledge, our work is the first to achieve 3D occupancy estimation on this dataset, thanks to our fully self-supervised learning setting. We present visualization results in Figure \ref{fig:occresult}. As highlighted by the red rectangle, the sky region has a short-range depth value, but this does not appear in the rendered 3D occupancy estimation map thanks to the parameterized coordinate design of OccNeRF~\cite{occnerf}. 

% Thanks to our fully self-supervised learning setting, we are able to perform 3D occupancy estimation on the DDAD dataset \cite{ddad}. 

\noindent \textbf{Depth estimation:} We present a comparison of depth estimation results in Table \ref{tab:depth benckmark} for both the nuScenes and DDAD. In stage 1, GaussianOcc $\ddagger$ achieves top performance on the nuScenes dataset and delivers competitive results on the DDAD. It is important to note that methods such as SurroundDepth \cite{surrounddepth} and SA-FSM \cite{SA-FSM} rely on third-party module for the sparse depth supervision. Additionally, SA-FSM is preprint work and has not released the code. R3D3 \cite{r3d3} is a temporal offline refinement method that requires multi-frame optimization, as discussed in \cite{selfocc}. 
In stage 2, which involves depth estimation from rendering, our method also achieves competitive results compared to those trained with ground truth poses. Besides, we observe that the rendered depth in stage two outperforms the depth results from stage 1 on the nuScenes, whereas the opposite is true for the DDAD. This discrepancy might be attributed to differences in perception range—80 meters in nuScenes versus 200 meters in DDAD. Moreover, an interesting phenomenon is that the semantic information is helpful for the depth estimation as indicated in GaussianOcc $\dagger$ whereas it worsens the result in OccNeRF $\dagger$. This phenomenon can be attributed to the biased sampling strategy of OccNeRF, where only 25\% of the sample points are used for faster semantic map rendering compared to depth map rendering. In contrast, our proposed Gaussian splatting method, which renders directly from the voxel vertices, eliminates this issue.

% We have a deeper analysis about this phenomenon in the supplement material.
% This shows the difference between the Gaussian splatting rendering and volume rendering.

\begin{figure}[t!]
\centering
{
\includegraphics[width=\linewidth]{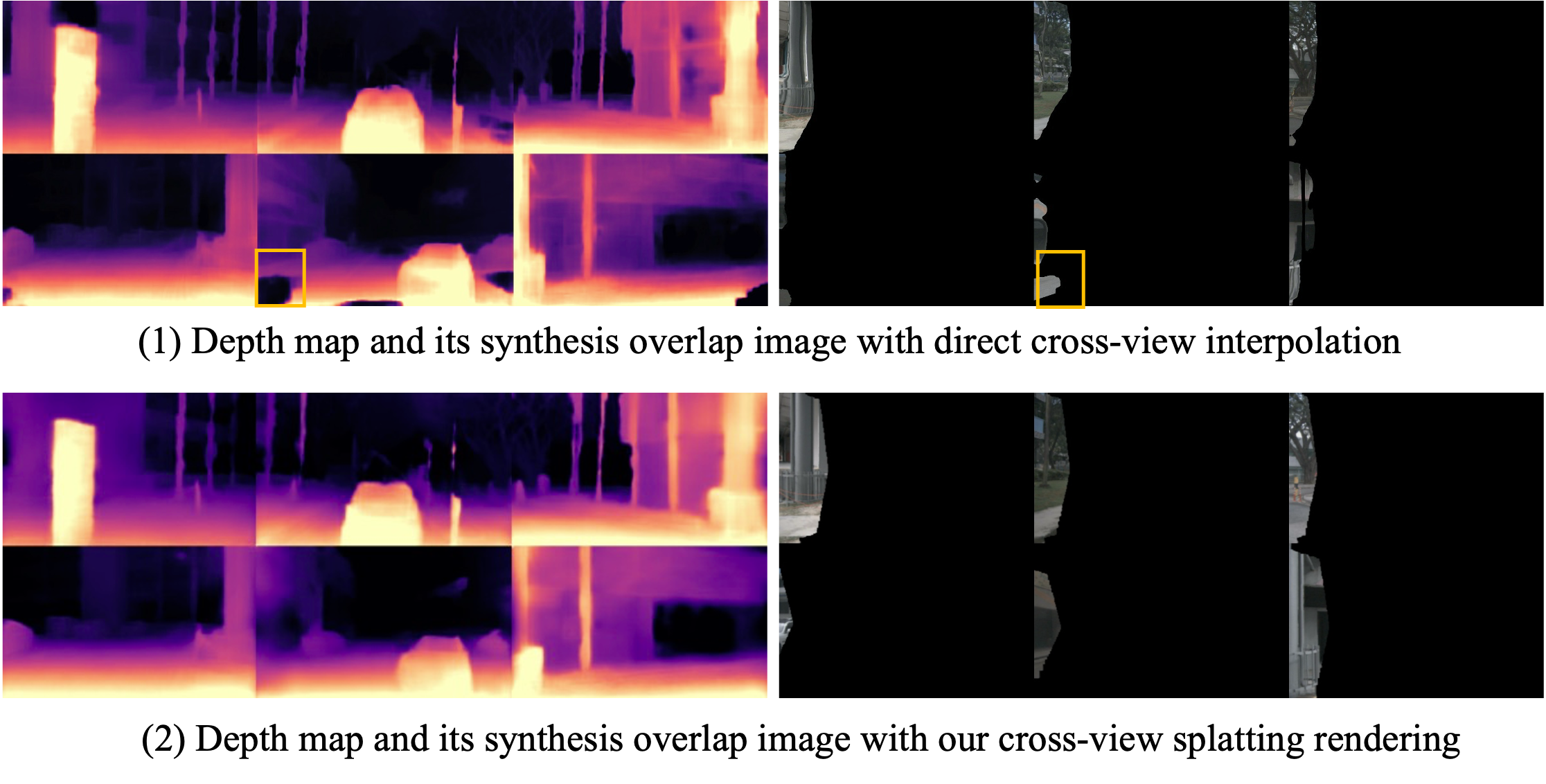}
}
 \vspace{-15pt}
\caption{The comparison of the depth map and its synthesis overlap image with (1) direct bilinear interpolation cross-view synthesis~\cite{surrounddepth} and (2) our cross-view Gaussian splatting synthesis.}
\label{fig:cross_view}
\end{figure}

\subsection{Ablation Study}

\noindent \textbf{Scale-aware training in Stage 1:} We compare our method with existing approaches \cite{surrounddepth, kim2022self} on Table \ref{tab:scale traning} to demonstrate the effectiveness of the proposed scale-aware training using Gaussian Splatting. For a more fair comparison, we also implemented scale-aware training on \cite{surrounddepth, kim2022self} that uses the same depth estimation network as ours \cite{newcrfs}, noted as $\checkmark$. We could reproduce the result on \cite{kim2022self}, but it is not good to training network with the sparse depth following \cite{surrounddepth}. Our proposed scale-aware training method is better than \cite{surrounddepth, kim2022self}. Specifically, we observed that a naive implementation of Gaussian splatting without a mask-out strategy for cross-view rendering is ineffective because it would lead to a sub-optimal solution that the rendered image is still from the current view. Besides, to enhance performance, we introduced an erosion operation on the binary mask to purify it, excluding regions that may fall outside the overlap area. This step ensures better alignment during training. Finally, we refined the depth estimation by fixing the 6D pose net and disabling the cross-view loss. This refinement helps reduce artifacts at the edges of the overlap region. \textbf{Visualization:} In Figure \ref{fig:cross_view}, we can see that the depth map is easily be suboptimal depth estimation in near overlap region with the training of (1) direct cross-view interpolation constrain in~\cite{surrounddepth}. Specifically, though the region with yellow rectangle has the incorrect large depth value, the synthesis region still presents a reasonable result that would lead to sub-optimal training. Thanks to our cross-view Gaussian splatting design analysis above, our method shown in (2) does not have the sub-optimal situation.

% We present the cross-view splatting render result in Figure \ref{fig:cross_view}, where the rendering result in overlap regions is reasonable in the correct depth estimation from cross view that verifies our design. 

\begin{figure}[t!]
\centering
{
\includegraphics[width=\linewidth]{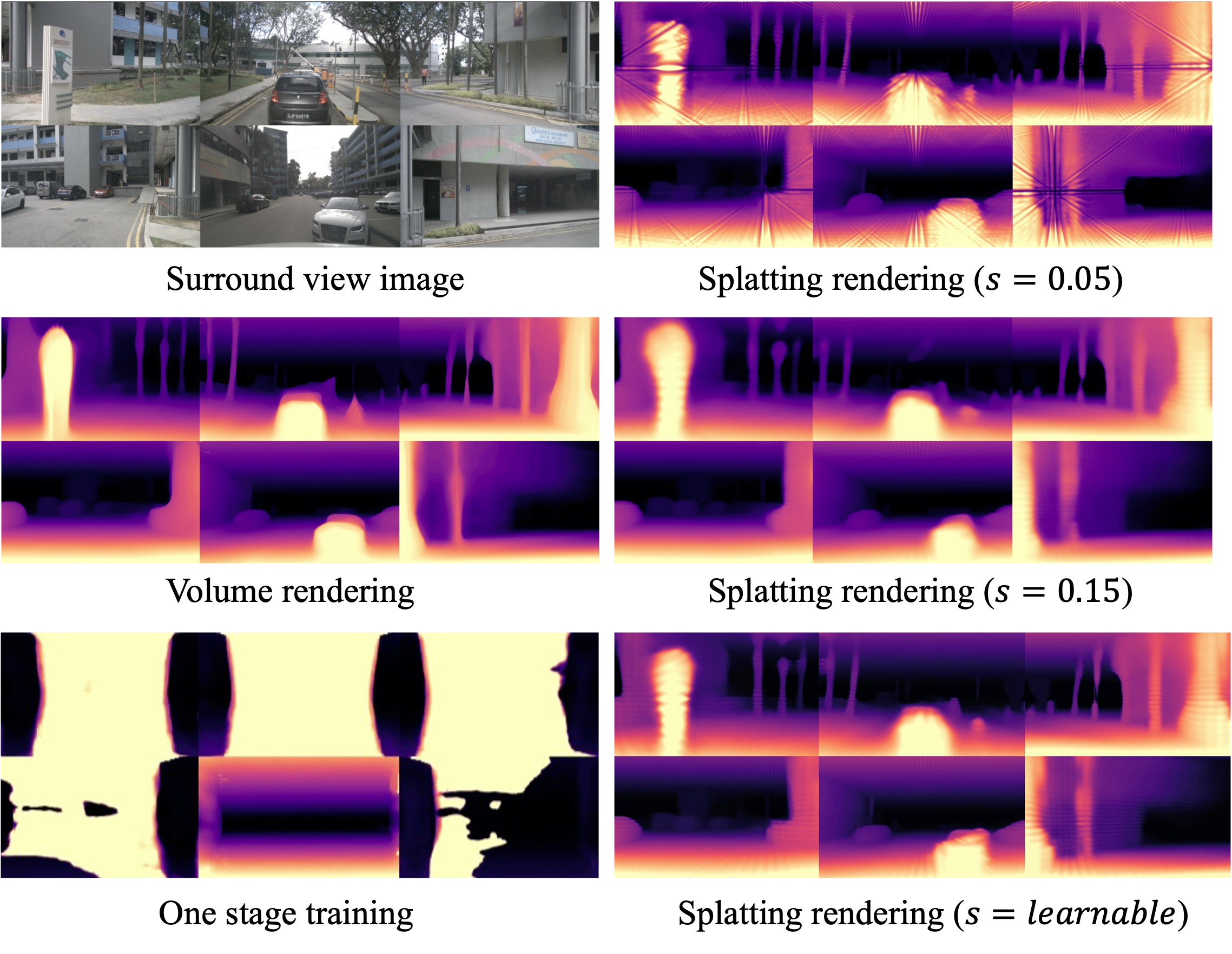}
}
 \vspace{-15pt}
\caption{The comparison for the depth map in the different setting, corresponding to the training strategy in Table \ref{tab:pose compare} and rendering type in Table \ref{tab:render type}.}
\label{fig:gs_scale}
\end{figure}

\begin{table}[t]
\centering
\scalebox{0.90}{
\begin{tabular}{c|c|c|c|c}
			\hline
			Pose type &\cellcolor{col1}Abs Rel & \cellcolor{col1}Sq Rel & \cellcolor{col1}RMSE &  \cellcolor{col2}$\delta < 1.25 $ \\
			
			\hline
% 0.214  &   3.362  &   7.127  &   0.321  &   0.771  &   0.881  &   0.932 
			 GT pose   &  0.214  &   3.362  &   \bf{7.127} &  \bf{0.771}
\\
% 0.211  &   3.115  &   7.131  &   0.326  &   0.762  &   0.878  &   0.931
          % 0.946  &  17.008  &  16.397  &   1.705  &   0.103  &   0.208  &   0.356 
              One stage training  & 0.946  &   17.008  &  16.397  & 0.103
\\

              \cite{kim2022self}   & 0.235  &   3.592  &   7.295 &  0.750
\\

             Ours   & \bf{0.211}  &   \bf{3.115}  &   7.131  &  0.762  
\\
\hline
	\end{tabular}}
	% \vspace{5pt}
	\caption{Comparison of pose type for stage 2 training on depth estimation task \cite{nuscenes}. One stage training directly uses the cross-view loss to the rendered depth map. Apart from the GT pose, we also experiment the learned pose from \cite{kim2022self} for comparison.}
	\label{tab:pose compare}
\end{table}

\begin{table}[t]
\centering
\scalebox{0.90}{
\begin{tabular}{c|c|c|c|c}
			\hline
			Render type &\cellcolor{col1}Abs Rel & \cellcolor{col1}Sq Rel & \cellcolor{col1}RMSE &  \cellcolor{col2}$\delta < 1.25 $ \\
		\hline		
% 0.215  &   3.508  &   7.113  &   0.324  &   0.775  &   0.881  &   0.931
			 VR   &  \underline{0.215}  &   3.508  &   \underline{7.113} &   \bf{0.775}
\\
\hline	
% 0.233  &   3.694  &   7.146  &   0.326  &   0.761  &   0.876  &   0.929
             SR ($s$ = 0.05)   &  0.223  &  3.694  &   7.246 &  0.761
\\
% 0.217  &   3.504  &   7.152  &   0.322  &   0.770  &   0.880  &   0.932
             SR ($s$ = 0.1)   &  0.217  &  3.504  &   7.152   &  0.770

\\
% 0.217  &   3.406  &   7.204  &   0.326  &   0.763  &   0.875  &   0.929
               SR ($s$ = 0.15)   & 0.217  &   \underline{3.406}  &   7.204  &  0.763 
\\
% % 0.212  &   3.248  &   7.112  &   0.321  &   0.771  &   0.879  &   0.931
               SR ($s$ = \textit{learnable})   & \textbf{0.212}  &   \textbf{3.248}  &   \textbf{7.112}  &  \underline{0.771}
\\
			\hline
	\end{tabular}}
	% \vspace{5pt}
	\caption{Comparison of the render result between the volume rendering (VR) \cite{occnerf} and splatting rendering (SR, Ours) on depth estimation task \cite{nuscenes}. The coefficient $s$ means the scale of the 3D Gaussian and \textit{learnable} means the scale is learnable by the Sigmoid function. We use GT pose for this set of the ablation study.}
	\label{tab:render type}
\end{table}

\begin{table}[t]
\centering
\resizebox{0.48\textwidth}{!}{
\begin{tabular}{c|c|c|c|c}
			\hline
			Pose type & mIoU* & \cellcolor{col1}Abs Rel & \cellcolor{col1}Sq Rel & \cellcolor{col2}$\delta < 1.25 $ \\
			
			\hline
% 0.456  &  12.682  &   9.194  &   0.399  &   0.704  &   0.833  &   0.890  
			 GT pose (VR)  & 10.81 &0.456  &  12.682 & 0.704
\\
% 0.225  &   4.339  &   6.568  &   0.303  &   0.787  &   0.895  &   0.940 
			 GT pose (SR)  & 11.30 &0.225  &   4.339 &   \bf{0.787}
\\
% 0.506  &  15.577  &   9.821  &   0.416  &   0.684  &   0.818  &   0.881  
             Learned pose (VR)  & 11.19 & 0.506  &  15.577 &  0.684
\\
%              Learned pose \cite{kim2022self}(SR)  & 11.19 & 0.506  &  15.577 &  0.684
% \\
 % \bf{0.197}  &   \bf{1.846}  &   6.733  &   \bf{0.312}  &   0.746  &   0.873  &   0.931       
              Learned pose (SR)  &  \bf{11.26} &  \bf{0.197}  &    \bf{1.846} &  0.746
\\
\hline
	\end{tabular}
 }
	% \vspace{5pt}
	\caption{Comparison of ground truth pose (GT Pose) and our learned pose (Two stages) on 3D occupancy estimation task  \cite{nuscenes} in volume rendering (VR) and splatting rendering (SR). 
	}
 \vspace{-5pt}
	\label{tab:pose_render}
\end{table}

\begin{table}[t]
	\centering
	\resizebox{0.48\textwidth}{!}{
		\begin{tabular}{c|c|c|c|c}
			\hline
			Render &  \multicolumn{3}{c|}{Render resolution and time (s)} & Training \\
     % \hline
                type &  180 $\times$ 320 & 240 $\times$ 520 & 360 $\times$ 640  &  time (h) \\
	        \hline
               VR & $\approx$ 0.85 &  $\approx$ 1.57 & N/A & $\approx$ 2.68 \\
    		\hline
    			SR & $\approx$ 0.17 & $\approx$ 0.17 & $\approx$ 0.17 & $\approx$ 1  \\
    		\hline
        \end{tabular}
    }
	\caption{Comparison of rendering efficiency between volume rendering (VR) \cite{occnerf} and splatting rendering (SR, Ours) on 3D occupancy estimation task~\cite{nuscenes}. The render time is calculated from surround 6 images. "N/A" indicates out-of-memory errors running in NVIDIA A 100 (40 GB). Training time is averaged per epoch.}

	\label{tab:rendertime}
\end{table}

% %%%%%%%%%%%%%%%%%%%%%%For %%%%%%%%%%%%%%%%%%%%%%
% \begin{table}[t]
% \centering
% \scalebox{0.90}{
% \begin{tabular}{c|c|c}

% \hline
%  Method & Pretraining setting  &  mIoU 
 
% \tabularnewline
% \hline

% Baseline &  None  &  37.29	 

% \tabularnewline
% \hline

% \multirow{2}*{Self-supervised pretraining} & DDAD  & 37.40 

% \tabularnewline

% ~ & nuScenes  &	\bf{38.45} 

% \tabularnewline
% \hline

% \end{tabular}}
% \caption{The study on SimpleOcc \cite{simpleocc} with fully self-supervised pretraining. The baseline is directly training the model with 3D occupancy labels. The self-supervised pretraining is conducted on DDAD and nuScenes and then finetuned the model with 3D occupancy label. The number with bold typeface means the best.}

% \vspace{0.3cm}
% \label{t:pretrain}
% \end{table}
% %%%%%%%%%%%%%%%%%%%%%% %%%%%%%%%%%%%%%%%%%%%%

\noindent \textbf{6D pose learning and training strategy:} Considering the 6D ego pose from the sensor is imperfect as a lack of vertical movement \cite{pose_issue}. We further evaluate the learned 6D pose quality by self-supervised depth result at stage 2, where the experiments on Table \ref{tab:pose compare} use the same network but with different pose constraints. Our result is competitive compared with GT pose, indicating that the predicted pose is of high quality. Furthermore, one stage training was ineffective in using the 6D pose from the jointly trained with 6D pose net because the cross-view loss in the 3D voxel space led to local optimization, which failed to generalize predictions to non-overlapping regions, as shown in Figure \ref{fig:gs_scale}. However, this issue did not occur in the depth maps produced by the 2D decoder in Stage 1, highlighting the necessity of the two-stage training. Besides, our result is also better than \cite{kim2022self}, which is consistent the depth result on Table \ref{tab:scale traning}.

% Additionally, the GT pose provided by the nuScenes dataset is limited by sensor inaccuracies, particularly a lack of vertical movement. In contrast, the learned 6D pose in our method does not suffer from this limitation. 

\noindent \textbf{Volume rendering and Splatting rendering:} 
In Table \ref{tab:render type}, we compare the performance of volume rendering (VR) and splatting rendering (SR) in the voxel space. For each 3D Gaussian, we first set a uniform scale \(s\) for all Gaussians, considering the well-arranged positions of vertices within the voxel grid. Since the scale \(s\) is consistent across all three dimensions, we set the rotation $\mathbf{R}$ as the identity matrix.
We found that a scale of 0.1 produced the best results, closely matching the VR at this scale. Besides, by applying a learnable scale through the output of a Sigmoid function and clamping the maximum scale to 0.12, we achieved the highest performance. The render depth maps are shown in Figure \ref{fig:gs_scale}. The depth map by the volume rendering is smoother thanks to dense sampling and the depth maps from splatting rendering have the graininess effect, especially at the small scale factor (0.05).

\noindent \textbf{Pose and render types ablation study in occupancy task:} In Table \ref{tab:pose_render}, we present an ablation study to evaluate the impact of using ground truth versus learned poses in the occupancy task, with training conducted using different rendering methods. The results show that splatting rendering achieves superior performance in both occupancy metrics (mIoU*) and depth metrics. One observation is that depth estimation results with the semantic learning in volume rendering are significantly worse than those in Table \ref{tab:depth benckmark} whereas our proposed splatting rendering method maintains consistent performance, which suggests the splatting rendering in voxel grid contributes better 3D geometry.

\noindent \textbf{Rendering efficiency analysis:} We analyze rendering efficiency in Table \ref{tab:rendertime}. The results show that volume rendering consumes 5 times more rendering time compared to splatting rendering. As resolution increases, both volume rendering time and GPU consumption rise significantly. In contrast, splatting rendering shows no significant increase in computational cost with higher resolutions, highlighting its efficiency and scalability.

% \noindent \textbf{Bonus of the fully self-supervised setting:} The fully self-supervised setting of our method could be a general pretraining solution for supervised learning. After the self-supervised training on the DDAD and nuScenes datasets, we further finetune the model with the 3D occupancy label from Occ3D~\cite{occ3d}. As shown in Table \ref{t:pretrain}, experiments with self-supervised pretraining outperform the baseline. In particular, we find that pretraining on nuScenes is better than the DDAD dataset, which may own to the domain gap factors, such as differences in the scenarios (RGB images) and sensor configurations (camera extrinsics). 

\section{Conclusion}
In this paper, we introduce GaussianOcc, a fully self-supervised and efficient method for 3D occupancy estimation. Through the carefully designed cross-view splatting rendering, we can accurately learn the real scale in depth and the 6D pose, enabling effective self-supervised 3D occupancy learning. Additionally, the proposed Gaussian splatting in voxel grids outperforms volume rendering in 3D occupancy estimation while reducing computational cost. 
% We validate our method on the nuScenes and DDAD datasets, demonstrating its practical significance.

% As a bonus, the fully self-supervised setting of the proposed method could serve as the general pretrain solution to enhance the 3D occupancy estimation performance. 

\textbf{Acknowledgments} This work was supported in part by the JSPS, KAKENHI under Grant Number 22H03609, JST, FOREST under Grant Number JPMJFR206S. Wanshui Gan was also supported by RIKEN JRA Program. Big thanks for Xiaoyu Dong for helping proofreading the manuscript. 

\newpage

\section*{Appendix}

\appendix

% \clearpage
% \setcounter{page}{1}
% \maketitlesupplementary

\begin{abstract}
In this supplementary material, we provide more implementation details, experiment results with analysis, and further discussion on the limitations and future work.
% Besides, we attach the main code and will release it to the public after the review.
\end{abstract}

%%%%%%%%% BODY TEXT
\section{More implementation details}

\noindent \textbf{The detailed parameter setting in Gaussian attributes estimation network \cite{newcrfs}.} 
During the joint depth and 6D pose training in stage 1, we predict the 3D Gaussian parameters alongside the 2D depth map. Since the Gaussian parameters are well-arranged in the 2D image plane prior to unprojection, we maintain equal scaling across all three dimensions of each 3D Gaussian and constrain the maximum scale to 0.02. Given that the scale \(s\) is uniform across all dimensions, we set the rotation matrix \(\mathbf{R}\) to the identity matrix. Additionally, we assign an opacity value of 1 to each 3D Gaussian, ensuring that every 2D depth value corresponds to a valid point in 3D space. We do not predict the color defined by SH coefficients $c$, while we directly use the source RGB image as the color map the same as in \cite{zheng2024gps}.

\textbf{The detailed parameter setting in voxel grid splatting rendering for semantic rendering.} For semantic rendering, we chose a fixed scale for each grid vertex to ensure a well-arranged structure that accurately models the 3D space. If we use a learnable scale, it may lead to a situation where the scale is small but the opacity is large, which may not be captured in the rendered depth map and semantic map but could still affect the 3D occupancy result. Therefore, using a fixed scale is simple and sufficient for optimization, as demonstrated in the results presented in the main paper (Table 5) that the performance is close to the learnable scale. Since the scale $s \in \mathbb{R}^3_+$ are identical for both three dimensions, we do not need to predict the rotation $r \in \mathbb{R}^4$ and set it with the identical matrix is sufficient. Similar to the OccNeRF \cite{occnerf}, we render the 2D feature map for the semantic regression, while we leverage the 3D Gaussian splatting rendering and OccNeRF uses volume rendering.

\textbf{The detailed parameter setting in training.} We follow the training setting as OccNeRF \cite{occnerf}, the resolution of input images and rendered depth maps are set as 384$\times$640 and 180$\times$320 respectively. All experiments are conducted on 8 NVIDIA A100 (40 GB).

\textbf{The detailed training strategy of the self-supervised pretraining setting.} We first do the self-supervised training with 12 epochs with the learned pose from stage 1 and the 2D pseudo semantic label, which does not require the 3D occupancy label. Then, we finetune the model with 12 epochs with the 3D occupancy label. We add the RayIoU metric \cite{sparseocc} in Table \ref{t:pretrain} for a comprehensive comparison.

\textbf{The detailed definition of depth map metric.} Following the depth estimation task \cite{surrounddepth}, we report the depth map evaluation with the following metrics, 
\begin{equation}
\begin{split}
\text {Abs Rel:} \frac{1}{|M|} \sum_{d \in M}\left|\hat{d}-d^*\right| / d^* , \\ 
\text {Sq Rel:} \frac{1}{|M|} \sum_{d \in M}\left\|\hat{d}-d^*\right\|^2 / d^*, \\
\text { RMSE: } \sqrt{\frac{1}{|M|} \sum_{d \in M}|| \hat{d}-d^* \|^2}, \\
\text { RMSE log: } \sqrt{\frac{1}{|M|} \sum_{d \in M}|| \log \hat{d}-\log d^* \|^2} , \\
\delta<t: \% \text { of } d \text { s.t. } \max \left(\frac{\hat{d}}{d^*}, \frac{d^*}{\hat{d}}\right)=\delta<t,
\end{split}
\end{equation}
where $M$ is the valid pixel, $\hat{d}$ is the ground truth depth and $d^*$ is the predicted depth.

\begin{figure}[t!]
\centering
{
\includegraphics[width=0.90\linewidth]{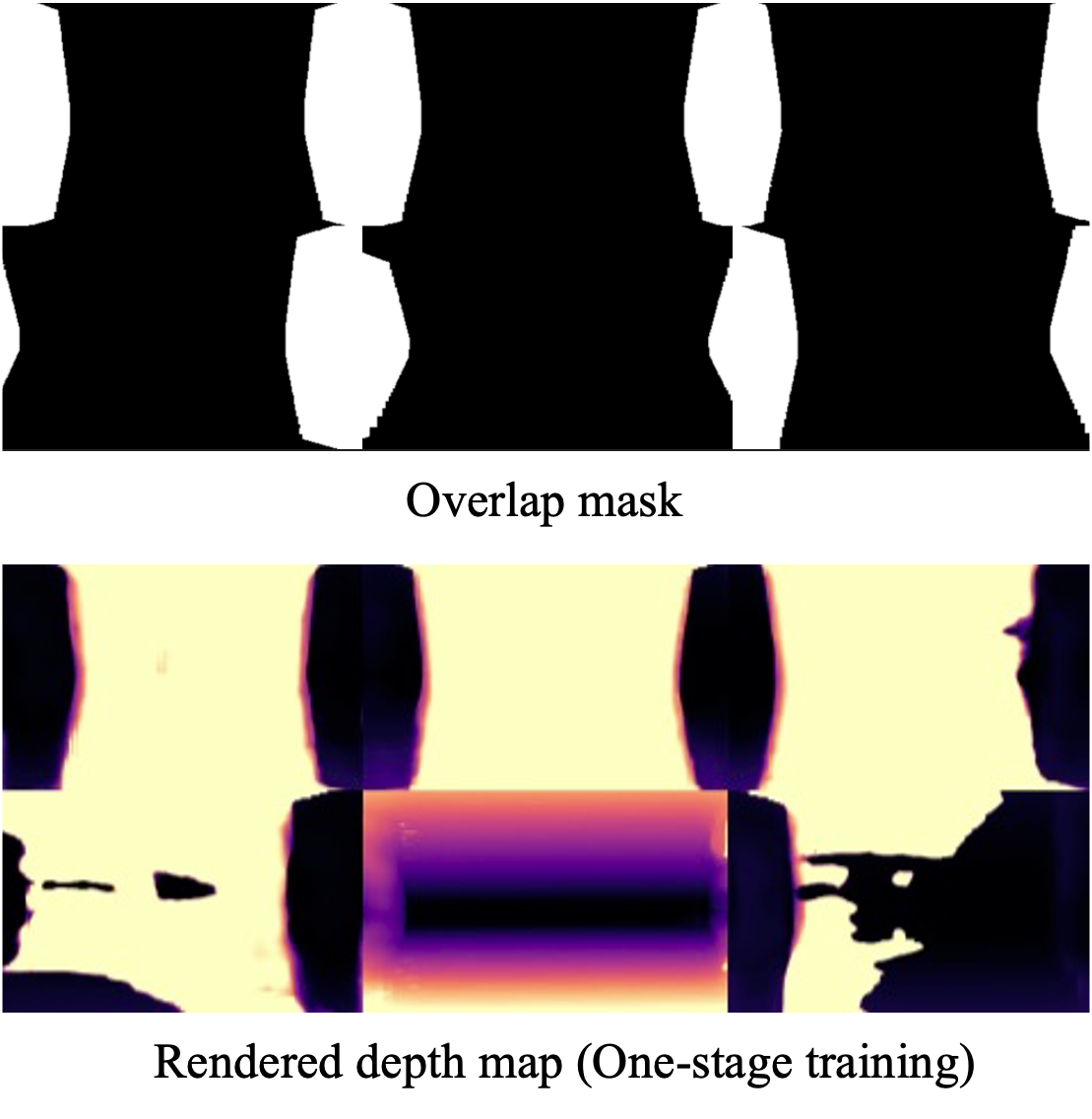}
}
 % \vspace{-25pt}
\caption{\textbf{One-stage training analysis}. This is the visualization of the overlap mask and the rendered depth map with one-stage training.}
\label{fig:onestage}
\end{figure}

\section{More experiment results and analysis}

\textbf{Why we need two-stage training.} We made extensive efforts to develop one-stage training that directly applies the cross-view loss to the rendered depth map but were unsuccessful. As suggested in Figure \ref{fig:onestage}, the cross-view supervision signals are effective only in overlapping regions. The rendered depth map learned from 3D CNN has lower generalization ability in non-overlap regions compared with the decoder depth learned by the 2D CNN, which led to local minima.

\textbf{mIoU metric.} Due to the limited space in the main paper, the full table of mIoU results is presented in Table \ref{tab:occ_supplement} for reference. 

\textbf{RayIoU metric.}  In addition to the mIoU metric for 3D occupancy estimation, we also evaluate our method with a novel metric, RayIoU, introduced by the recent work \cite{sparseocc}. The RayIoU is a ray-based evaluation metric that resolves the inconsistency penalty along the depth axis introduced in the traditional voxel-level mIoU criteria. As shown in Table \ref{table:rayiou}, our approach also outperforms OccNeRF \cite{occnerf} in this metric as well. It's important to note that the FPS is calculated excluding rendering time. Since GaussianOcc and OccNeRF utilize the same network architecture, they share the same inference time when the rendering process is not taken into account.

% \textbf{More visualization.} We provide more visualization for nuScenes dataset in Figure \ref{fig:nuscenes_supplement} and DDAD dataset in Figure \ref{fig:ddad_supplement}. Please check the attached videos for sequence visualization.

\begin{figure*}
    \centering
    \includegraphics[width=0.90 \textwidth]{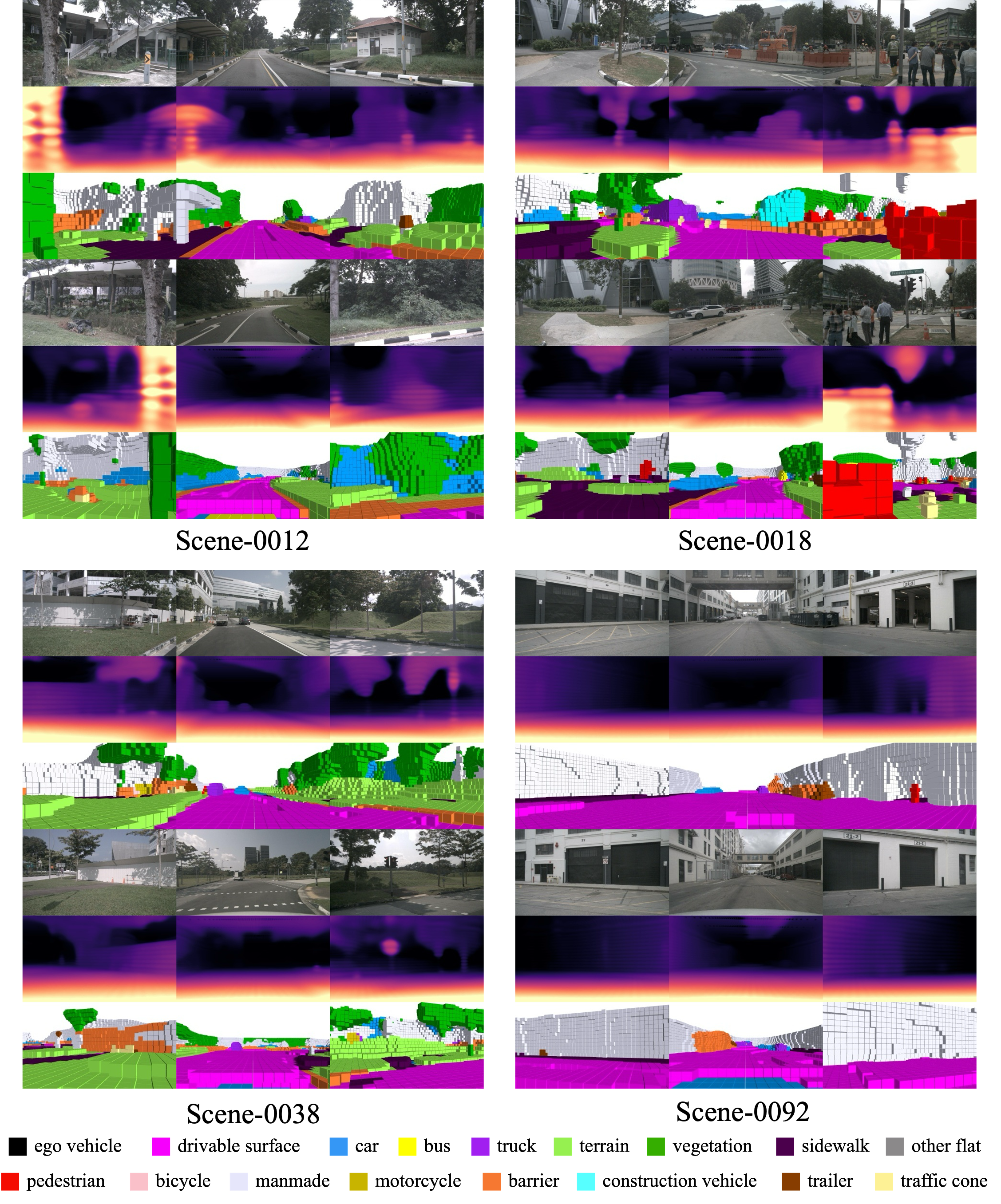}
     \vspace{-10pt}
    \caption{\textbf{The visualization of the render depth map and 3D occupancy prediction on nuScenes dataset.} }
    \label{fig:nuscenes_supplement}
\end{figure*}

\textbf{More visualization.} We provide more visualization for nuScenes dataset in Figure \ref{fig:nuscenes_supplement}. Please check the videos for sequence visualization in \href{https://github.com/GANWANSHUI/GaussianOcc.git}{https://github.com/GANWANSHUI/GaussianOcc.git}.

\textbf{More analysis on 3D occupancy and depth map result on different supervision types.} 

\begin{figure*}
    \centering
    \includegraphics[width=0.90 \textwidth]{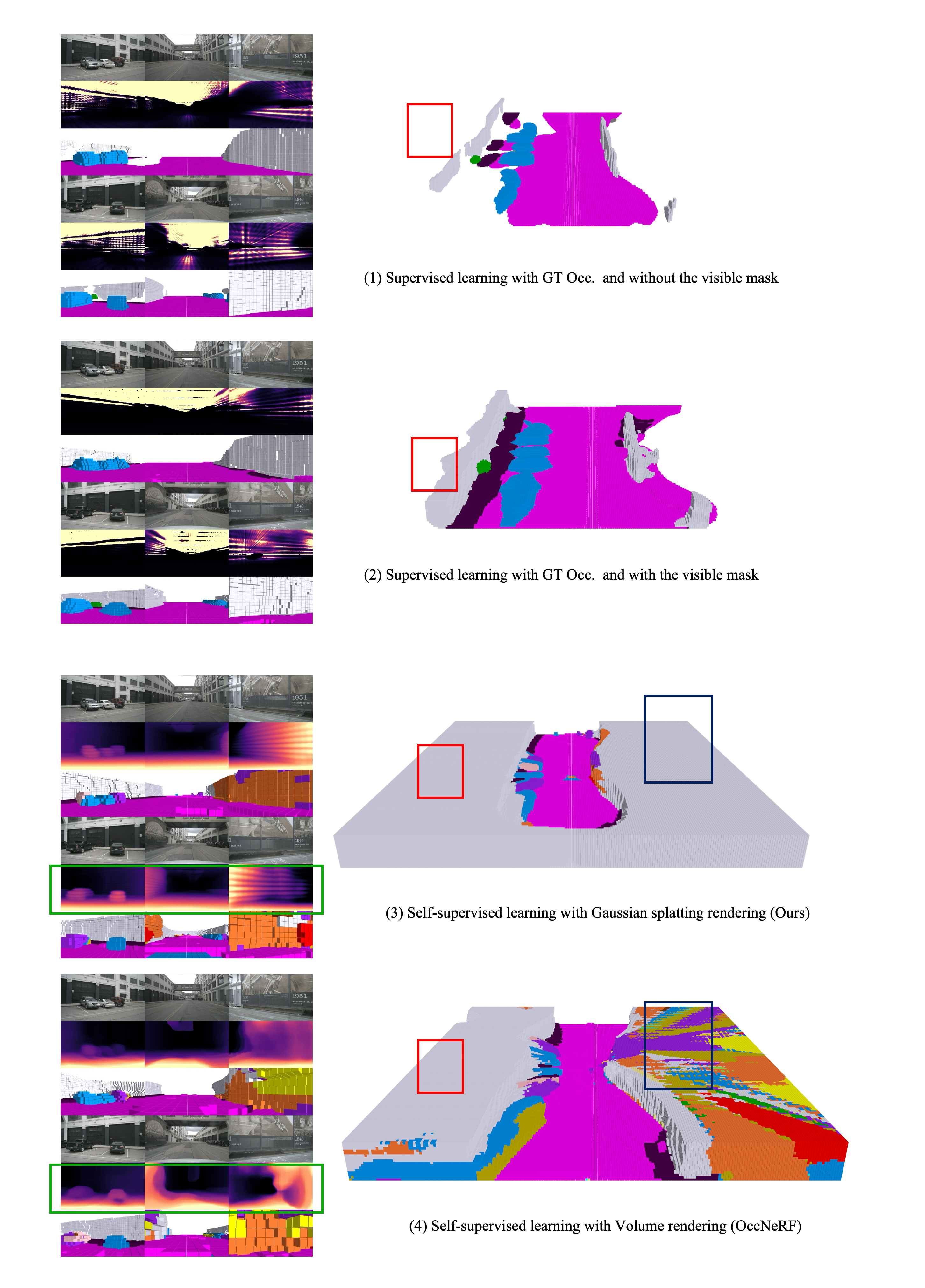}
     \vspace{-10pt}
    \caption{\textbf{The visualization of the different supervision types (1-4) comparison on nuScenes dataset.} }
    \label{fig:supervision_compare}
\end{figure*}

\noindent \textbf{3D occupancy analysis:} In Figure \ref{fig:supervision_compare}, we present visualizations of different supervision types. These visualizations highlight key differences in the results for the invisible regions (marked with red rectangles) and the rendered depth quality (marked with green rectangles).

Experiments (1) and (2) involve supervision using ground truth (GT) occupancy labels. Specifically:

Experiment (1) is trained without the visible mask provided by Occ3D-nuScenes \cite{occ3d}, which defines the visibility of the occupancy labels. Without this mask, the invisible regions are treated as empty, and the loss function is applied to these regions as well.
Experiment (2), on the other hand, excludes the loss computation in invisible regions.
From the results, we observe that in Experiment (1), the model tends to predict empty values for invisible regions due to empty loss penalty. In contrast, Experiment (2), by ignoring the loss in invisible areas, shows more non-empty predictions in these regions.

Self-supervised experiments (3) and (4) rely on rendering techniques, which inherently cannot optimize predictions in invisible regions. This limitation leads to non-empty predictions in the red-highlighted areas. Notably, Experiment (4) frequently predicts invisible regions as related to foreground categories, as shown in the dark rectangles. Conversely, Experiment (3) demonstrates a consistent tendency to classify invisible regions as man-made structures, likely because the surrounding environment predominantly consists of man-made elements.

\noindent \textbf{Render depth map analysis:} In Tables 2 and 6 of main paper, we observe an interesting phenomenon that the semantic information is helpful for the depth estimation with our GaussianOcc whereas it worsens the result in OccNeRF. In Figure \ref{fig:supervision_compare}, we visualize the depth map and highlight with green rectangles that our Gaussian splatting rendering produces higher-quality depth predictions compared to volume rendering. This should be concluded to the biased sampling strategy of OccNeRF, where only 25\% of the sample points are used for faster semantic map rendering compared to depth map rendering. Here is the piece of the \href{https://github.com/LinShan-Bin/OccNeRF/blob/dd67c728aa19b50a936b2f8ec9c9e56dc9404fc7/options.py#L81}{code} in OccNeRF~\cite{occnerf}. In contrast, our proposed Gaussian splatting method, which renders directly from the voxel vertices, eliminates this issue. At last, since Experiments (1) and (2) do not involve rendering-based training, they fail to produce reasonable depth predictions.

\begin{table}[t]
	\centering
	\resizebox{0.48\textwidth}{!}{
		\begin{tabular}{c|c|c|c}
		\hline
		Render resolution  &  \multicolumn{3}{c}{ Render time (s) with different voxel resolutions (Gaussians number)}  \\

          $180 \times 320$  &  $16 \times 200 \times 200$  &  $24 \times 300 \times 300$  & $32 \times 512 \times 512$  \\
	\hline
    			VR & $\approx$ 0.50 & $\approx$ 0.85 & $\approx$ 1.52   \\
            \hline
    			SR & $\approx$ 0.06 & $\approx$ 0.17 & $\approx$ 0.44  \\
    		\hline
        \end{tabular}
    }
\caption{Comparison of rendering efficiency under different Gaussians number between volume rendering (VR) \cite{occnerf} and splatting rendering (SR, Ours).}
\label{t:gaussian_number}    
\end{table}

\noindent \textbf{Gaussians number and its related render time:} (1) In stage~1, Gaussians number depends on the depth map resolution from the 2D decoder, where each pixel is a Gaussian primitive after unprojection. We use the depth map resolution in $224 \times 352$, resulting in 78,848 Gaussian primitives in one image. In stage 2, Gaussians number depends on the voxel resolution, where each voxel grid is a Gaussian primitive. In Table 7 of the main paper, we follow the voxel resolution the same as OccNeRF [52] in $24 \times 300 \times 300$, resulting in 2,160,000 Gaussian primitives. (2) We revealed the rendering time under different render image resolutions compared with volume rendering in Table 7 of the main paper. We conducted the extra experiment for render time comparison under the same render image resolution ($180 \times 320$) but with different Gaussians number (voxel resolutions) as shown in Table \ref{t:gaussian_number}. From Table 7 of the main paper and Table \ref{t:gaussian_number}, we observe: (1) The render time of splatting rendering (SR) is mainly affected by the Gaussians number, not the render image resolution. (2) SR is 3–8 times faster than volume rendering (VR) across different voxel settings.

% %%%%%%%%%%%%%%%%%%%%%%For %%%%%%%%%%%%%%%%%%%%%%
% \begin{table}[t]
% \centering
% \scalebox{0.90}{
% \begin{tabular}{c|c|c}

% \hline
%  Method & Pretraining setting  &  mIoU 
 
% \tabularnewline
% \hline

% Baseline &  None  &  37.29	 

% \tabularnewline
% \hline

% \multirow{2}*{Self-supervised pretraining} & DDAD  & 37.40 

% \tabularnewline

% ~ & nuScenes  &	\bf{38.45} 

% \tabularnewline
% \hline

% \end{tabular}}
% \caption{The study on SimpleOcc \cite{simpleocc} with fully self-supervised pretraining. The baseline is directly training the model with 3D occupancy labels. The self-supervised pretraining is conducted on DDAD and nuScenes and then finetuned the model with 3D occupancy label. The number with bold typeface means the best.}

% \vspace{0.3cm}
% \label{t:pretrain}
% \end{table}
% %%%%%%%%%%%%%%%%%%%%%% %%%%%%%%%%%%%%%%%%%%%%

\noindent \textbf{Bonus of the fully self-supervised setting:} The fully self-supervised setting of our method could be a general pretraining solution for supervised learning. After the self-supervised training on the DDAD and nuScenes datasets, we further finetune the model with the 3D occupancy label from Occ3D~\cite{occ3d}. As shown in Table \ref{t:pretrain}, experiments with self-supervised pretraining outperform the baseline. In particular, we find that pretraining on nuScenes is better than the DDAD dataset, which may own to the domain gap factors, such as differences in the scenarios (RGB images) and sensor configurations (camera extrinsics).

\section{Limitation and future work}

\begin{figure*}
    \centering
    \includegraphics[width=0.90 \textwidth]{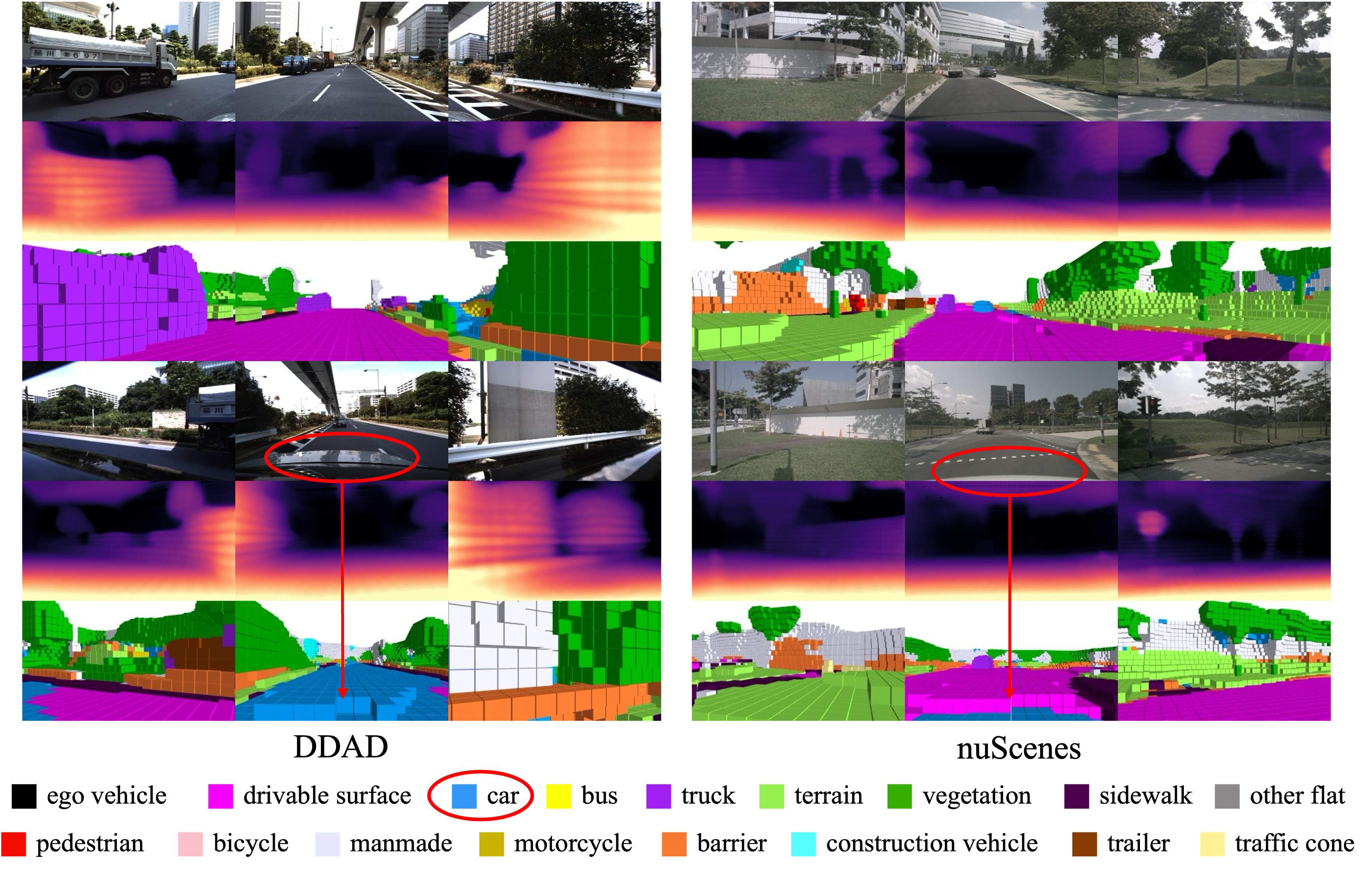}
     \vspace{-10pt}
    \caption{\textbf{Some wrong predictions due to the large self-occlusion on DDAD dataset.} }
    \label{fig:limitation}
\end{figure*}

The proposed method achieves reasonable predictions in most scenes; however, we observe that some cases still present challenges, as shown in Figure \ref{fig:limitation}. Specifically, in the DDAD dataset, incorrect predictions occur in the back camera in certain situations as marked with the red circle, where the drivable surface is mistakenly projected into the car due to extensive self-occlusion. Notably, this issue is absent in the nuScenes dataset, which has less self-occlusion. We believe that this problem could be mitigated with better 2D semantic maps for supervision, which warrants further investigation. The proposed method is for the surround view setting which is not suitable for the monocular images. Additionally, in stage 1, we leverage the spatial cross-view constraint for scale-aware training through the proposed Gaussian splatting method. In the future, we aim to explore its potential benefits for temporal view synthesis as well.

% we plan to explore the use of splatting rendering for 3D occupancy flow.

\definecolor{nbarrier}{RGB}{255, 120, 50}
\definecolor{nbicycle}{RGB}{255, 192, 203}
\definecolor{nbus}{RGB}{255, 255, 0}
\definecolor{ncar}{RGB}{0, 150, 245}
\definecolor{nconstruct}{RGB}{0, 255, 255}
\definecolor{nmotor}{RGB}{200, 180, 0}
\definecolor{npedestrian}{RGB}{255, 0, 0}
\definecolor{ntraffic}{RGB}{255, 240, 150}
\definecolor{ntrailer}{RGB}{135, 60, 0}
\definecolor{ntruck}{RGB}{160, 32, 240}
\definecolor{ndriveable}{RGB}{255, 0, 255}
\definecolor{nother}{RGB}{139, 137, 137}
\definecolor{nsidewalk}{RGB}{75, 0, 75}
\definecolor{nterrain}{RGB}{150, 240, 80}
\definecolor{nmanmade}{RGB}{230, 230, 250}
\definecolor{nvegetation}{RGB}{0, 175, 0}
\definecolor{nothers}{RGB}{0, 0, 0}

\definecolor{col1}{RGB}{232, 161, 148}
\definecolor{col2}{RGB}{148, 187, 232}

\begin{table*}[ht]
	\footnotesize
 	\setlength{\tabcolsep}{0.0025\linewidth}
	
	\newcommand{\classfreq}[1]{{~\tiny(\nuscenesfreq{#1}\%)}}  %
    \begin{center}
	\resizebox{1.0\textwidth}{!}{
	\begin{tabular}{l|c c|c c| c c c c c c c c c c c c c c c}
		\toprule
		Method
		& \rotatebox{90}{GT Occ.} & \rotatebox{90}{GT Pose} & mIoU* & mIoU
        
		& \rotatebox{90}{\textcolor{nbarrier}{$\blacksquare$} barrier}
		
		& \rotatebox{90}{\textcolor{nbicycle}{$\blacksquare$} bicycle}
		
		& \rotatebox{90}{\textcolor{nbus}{$\blacksquare$} bus}

		& \rotatebox{90}{\textcolor{ncar}{$\blacksquare$} car}

		& \rotatebox{90}{\textcolor{nconstruct}{$\blacksquare$} const. veh.}

		& \rotatebox{90}{\textcolor{nmotor}{$\blacksquare$} motorcycle}

		& \rotatebox{90}{\textcolor{npedestrian}{$\blacksquare$} pedestrian}

		& \rotatebox{90}{\textcolor{ntraffic}{$\blacksquare$} traffic cone}

		& \rotatebox{90}{\textcolor{ntrailer}{$\blacksquare$} trailer}

		& \rotatebox{90}{\textcolor{ntruck}{$\blacksquare$} truck}

		& \rotatebox{90}{\textcolor{ndriveable}{$\blacksquare$} drive. suf.}

		& \rotatebox{90}{\textcolor{nsidewalk}{$\blacksquare$} sidewalk}

		& \rotatebox{90}{\textcolor{nterrain}{$\blacksquare$} terrain}

		& \rotatebox{90}{\textcolor{nmanmade}{$\blacksquare$} manmade}

		& \rotatebox{90}{\textcolor{nvegetation}{$\blacksquare$} vegetation}
        
		\\
		\midrule
    MonoScene~\cite{monoscene} & \checkmark & $\times$ & 6.33 & 6.06  & 7.23 & 4.26 & 4.93 & 9.38 & 5.67 & 3.98 & 3.01 & 5.90 & 4.45 & 7.17 & 14.91  & 7.92 & 7.43 & 1.01 & 7.65  \\
    BEVDet ~\cite{bevdet} & \checkmark & $\times$ & 20.03 & 19.38 & 30.31 & 0.23 & 32.26 & 34.47 & 12.97 & 10.34 & 10.36 & 6.26 & 8.93 & 23.65 & 52.27 & 26.06 & 22.31 & 15.04 & 15.10  \\
    BEVFormer~\cite{bevformer} & \checkmark & $\times$ & 24.64 & 23.67& 38.79 &9.98 &34.41 &41.09 &13.24 &16.50& 18.15 &17.83 &18.66 &27.70 &48.95 &29.08 &25.38 &15.41 &14.46 \\
    OccFormer~\cite{occformer}& \checkmark & $\times$ & 22.39& 21.93 & 30.29 & 12.32 & 34.40 & 39.17 & 14.44 & 16.45 & 17.22 & 9.27 & 13.90 & 26.36 & 50.99 &  34.66 & 22.73 & 6.76 & 6.97  \\
    RenderOcc~\cite{renderocc} & \checkmark & $\times$ &  24.53 &23.93&27.56 &14.36 &19.91 &20.56 &11.96 &12.42 &12.14 &14.34 &20.81 &18.94 &68.85 & 42.01 &43.94 &17.36& 22.61 \\ 
    TPVFormer~\cite{tpvformer} &\checkmark & $\times$ & 28.69 & 27.83 & 38.90 &13.67 &40.78 &45.90 &17.23& 19.99& 18.85& 14.30& 26.69& 34.17& 55.65& 37.55& 30.70& 19.40& 16.78 \\
    CTF-Occ~\cite{occ3d} & \checkmark & $\times$ & \bf{29.54} & \bf{28.53} &39.33 &20.56 &38.29& 42.24 &16.93 &24.52 &22.72 &21.05 &22.98 &31.11& 53.33 & 37.98 &33.23 &20.79 &18.00
    \\
    
        \midrule
        SimpleOcc~\cite{simpleocc} & $\times$ & \checkmark  & 7.99 & 7.05 & 0.67 &1.18 &3.21 &7.63 &1.02& 0.26 &1.80 &0.26 &1.07 &2.81 &40.44 &18.30 &17.01 &13.42 &10.84 \\
         SelfOcc~\cite{selfocc} &  $\times$ & \checkmark & 10.54  & 9.30 & 0.15 &0.66 &5.46 & 12.54 & 0.00 & 0.80 & 2.10 
         & 0.00 & 0.00 & 8.25 & 55.49 & 26.30 & 26.54 & 14.22 & 5.60
         \\
         OccNeRF~\cite{occnerf} &  $\times$ & \checkmark & 10.81  & 9.54 & 0.83 & 0.82 & 5.13 & 12.49 & 3.50 & 0.23 & 3.10 & 1.84 & 0.52 & 3.90 & 52.62 & 20.81 & 24.75 & 18.45 & 13.19 \\

          GaussianOcc &  $\times$ & $\times$ & \bf{11.26}  & \bf{9.94} & 1.79 & 5.82 & 14.58 & 13.55 & 1.30 & 2.82 & 7.95 & 9.76 & 0.56 & 9.61 & 44.59 & 20.10 & 17.58 & 8.61 & 10.29 \\
          
	\bottomrule
	\end{tabular}}
    \end{center}
     \vspace{-12pt}
    \caption{\textbf{3D occupancy prediction performance on the Occ3D-nuScenes dataset in mIoU metric.} Since `other' and `other flat' classes are the invalid prompts for open-vocabulary models, we also calculate `mIoU*' as the result ignoring the classes that do not consider these two classes during evaluation, while `mIoU' is the original result. GT Occ. refers to the use of the ground truth occupancy label for supervision. GT Pose is the ground truth pose from the sensor for self-supervised geometry learning.}
	\label{tab:occ_supplement}

\end{table*}

\begin{table*}[t]
  \setlength{\tabcolsep}{5pt}
   \centering
   \scalebox{0.91}{
   \begin{tabular}{l|cc|ccc|c|ccc|c|c}
      \toprule
      Method & \rotatebox{90}{GT Occ.} & \rotatebox{90}{GT Pose} & Backbone & Input Size & Epoch & \cellcolor[gray]{0.93}{RayIoU} & \multicolumn{3}{c|}{RayIoU\textsubscript{1m, 2m, 4m}} & mIoU & FPS \\
      \midrule
      BEVFormer (4f) \cite{bevformer} & \checkmark & $\times$ & R101   & 1600$\times$900 & 24 & \cellcolor[gray]{0.93}{32.4} & 26.1 & 32.9 & 38.0 & 39.2 & 3.0 \\
      RenderOcc \cite{renderocc}  & \checkmark & $\times$ & Swin-B & 1408$\times$512 & 12 & \cellcolor[gray]{0.93}{19.5} & 13.4 & 19.6 & 25.5 & 24.4 & - \\
      SimpleOcc \cite{simpleocc}  & \checkmark & $\times$ & R101 & 672$\times$336 & 12 & \cellcolor[gray]{0.93}{28.2} & 22.3 & 28.7 & 33.7 & 37.3 & 9.7 \\
      BEVDet-Occ (2f) \cite{bevdet4d}  & \checkmark & $\times$ & R50    & 704$\times$256  & 90 & \cellcolor[gray]{0.93}{29.6} & 23.6 & 30.0 & 35.1 & 36.1 & 2.6 \\
      BEVDet-Occ-Long (8f)    & \checkmark & $\times$ & R50    & 704$\times$384  & 90 & \cellcolor[gray]{0.93}{32.6} & 26.6 & 33.1 & 38.2 & \textbf{39.3} & 0.8 \\
      FB-Occ (16f) \cite{fbocc}       & \checkmark & $\times$ & R50    & 704$\times$256  & 90 & \cellcolor[gray]{0.93}{33.5} & 26.7 & 34.1 & 39.7 & 39.1 & 10.3 \\
      % \midrule
      SparseOcc (8f)                 & \checkmark & $\times$ & R50    & 704$\times$256  & 24 & \cellcolor[gray]{0.93}{34.0} & 28.0 & 34.7 & 39.4 & 30.1 & \textbf{17.3} \\
      SparseOcc (16f)                & \checkmark & $\times$ & R50    & 704$\times$256  & 48 & \cellcolor[gray]{0.93}{\textbf{36.1}} & \textbf{30.2} & \textbf{36.8} & \textbf{41.2} & 30.9 & 12.5 \\
       \midrule

       OccNeRF \cite{occnerf}          & $\times$  & \checkmark & R101  & 640$\times$384  & 12 &  \cellcolor[gray]{0.93}{10.49} & 6.93 & 10.28 &  14.26 & 9.54 & 10.8 \\

        GaussianOcc   & $\times$  & $\times$ & R101  & 640$\times$384  & 12 & \cellcolor[gray]{0.93}{\textbf{11.85}} & \textbf{8.69} & \textbf{11.90} &  \textbf{14.95} & \textbf{9.94} & 10.8 \\
      
      \bottomrule
      
   \end{tabular}
   }
     \caption{\textbf{3D Occupancy prediction performance on the Occ3D-nuScenes dataset in RayIoU metric.} GT Occ. means using the ground truth occupancy label for the supervision. GT Pose is the ground truth pose from the sensor for self-supervised geometry learning. ``8f'' and ``16f'' mean fusing temporal information from 8 or 16 frames. mIoU is the mean Intersection over Union for all categories.  FPS means frame per second for each method, which is measured on a Tesla A100 GPU.}
     \label{table:rayiou}
\end{table*}

% \begin{figure*}
%     \centering
%     \includegraphics[width=0.90 \textwidth]{figure/ddad_supplement.png}
%      \vspace{-10pt}
%     \caption{\textbf{The visualization of the render depth map and 3D occupancy prediction on DDAD dataset} }
%     \label{fig:ddad_supplement}
% \end{figure*}

%%%%%%%%%%%%%%%%%%%%%%For %%%%%%%%%%%%%%%%%%%%%%
\begin{table*}[t]
\centering
\scalebox{0.85}{
\begin{tabular}{c|c|c|c c c c}

\hline
 Method & Pretraining setting  &  mIoU & \cellcolor[gray]{0.93}{RayIoU} & \multicolumn{3}{c}{RayIoU\textsubscript{1m, 2m, 4m}} 
        \\
    \midrule

Baseline &  None  &  37.29 & \cellcolor[gray]{0.93}{28.2} & 22.3 & 28.7 & 33.7 

\tabularnewline
\hline

\multirow{2}*{Self-supervised pretrain} & DDAD  & 37.40 & \cellcolor[gray]{0.93}{28.7} & 22.9 & 29.1 & 34.0 

\tabularnewline

~ & nuScenes  &	\bf{38.45} & \cellcolor[gray]{0.93}{\bf{29.9}} & \bf{23.9} & \bf{30.4} & \bf{35.5} 
\tabularnewline
\hline

% {'RayIoU': 0.29950570216486677, 'RayIoU@1': 0.23944308625391203, 'RayIoU@2': 0.3039886542146309, 'RayIoU@4': 0.35508536602605734}

\end{tabular}}
\caption{The study on SimpleOcc \cite{simpleocc} with fully self-supervised pretrain. The baseline is directly training the model with 3D occupancy label. The self-supervised pretraining is conducted on DDAD and nuScenes and then finetuned the model with 3D occupancy label. The number with bold typeface means the best.}

\vspace{0.3cm}
\label{t:pretrain}
\end{table*}
%%%%%%%%%%%%%%%%%%%%%% %%%%%%%%%%%%%%%%%%%%%%

{
    \small
    \bibliographystyle{ieeenat_fullname}
    \bibliography{main}
}

\end{document}